\documentclass[preprint,12pt]{elsarticle}

\usepackage{amssymb}
\usepackage{amsmath}

\journal{Artificial Intelligence Journal}

\usepackage{xeCJK}
\usepackage{xcolor}
\usepackage{amsmath}
\usepackage{booktabs}
\usepackage[tone]{tipa}
\usepackage[utf8]{inputenc}
\usepackage{makecell}
\usepackage{latexsym}
\usepackage{graphicx}
\usepackage{subcaption}
\usepackage{microtype}
\usepackage{inconsolata}
\usepackage{amssymb}
\usepackage{multirow}
\usepackage{etoolbox}
\usepackage{url}

\usepackage[ruled,vlined]{algorithm2e}
\SetAlCapSkip{3pt}

\begin{document}

\begin{frontmatter}



\title{Beyond Atomic Tokens: Factorizing Syllables for Language Model Pretraining}


\author{Nghia Hieu Nguyen\fnref{1}}
\ead{nghiangh@uit.edu.vn}

\author{Thai Bao Huynh\fnref{1}}
\ead{23520105@gm.uit.edu.vn}

\author{Binh-An Dinh-Le\fnref{1}}
\ead{23520004@gm.uit.edu.vn}

\author{Phu Gia Hoang\fnref{1}}
\ead{19520215@gm.uit.edu.vn}

\author{Dat Tien Nguyen\fnref{2}}
\ead{tiendat.nguyen@mbzuai.ac.ae}

\author{Kiet Van Nguyen\fnref{1}}
\ead{kietnv@uit.edu.vn}

\author{Ngan Luu-Thuy Nguyen\fnref{1}}
\ead{ngannlt@uit.edu.vn}

\affiliation[1]{organization={University of Information Technology, Vietnam National University},
            city={Ho Chi Minh city},
            country={Viet Nam}}

\affiliation[2]{organization={Mohamed bin Zayed University of Artificial Intelligence, Abu Dhabi, UAE},
            city={Abu Dhabi},
            country={UEA}}

\begin{abstract}
Conventional tokenizers represent text as characters or statistically derived subwords, overlooking the internal phonological structure of syllables and often requiring large vocabularies. We introduce \textbf{Phonemic Tokenizer}, a linguistically motivated tokenizer for Vietnamese and Chinese that converts each syllable into IPA and factorizes it into three phonological components: onset, rime, and tone. The three components jointly occupy one contextual position, preserving syllable-level sequence length while enabling representation sharing across phonologically related syllables. Non-phonological and unsupported units are handled through character-level fallback. This deterministic design requires no corpus-dependent vocabulary learning and yields vocabularies of only 112 entries for Chinese and 256 for Vietnamese. Intrinsic evaluation shows that the tokenizer achieves substantially higher Rényi efficiency in both languages, represents every entry in a standard Vietnamese syllable dictionary with a Fertility of exactly one, and generally produces shorter Vietnamese sequences than existing pretrained tokenizers. We further instantiate the tokenizer in \textbf{PhonemicBERT}, which combines factorized component embeddings and reconstructs complete masked syllables using three prediction heads. Under a controlled Chinese pretraining setup, PhonemicBERT-Zh is competitive with or outperforms character, subword, and SubChar alternatives across diverse language-understanding tasks. PhonemicBERT-Vi also achieves competitive or superior results to established Vietnamese and multilingual pretrained models. These results establish phonemic factorization as a compact, efficient, and interpretable alternative to atomic and statistically segmented text representations.
\end{abstract}


\begin{highlights}
\item A linguistically motivated tokenizer explicitly models phonological structure.
\item Compact compositional representations enable effective parameter sharing.
\item The proposed approach supports competitive and robust language understanding.
\end{highlights}


\end{frontmatter}



\section{Introduction}
\label{sec:introduction}

Tokenization determines the computational units through which pretrained language models (PLMs) access text. Most PLMs rely on corpus-induced algorithms such as Byte Pair Encoding (BPE) \cite{sennrich2016neural}, WordPiece \cite{schuster2012japanese}, or the Unigram Language Model \cite{kudo2018subword}. Although effective, these methods do not guarantee that the resulting tokens correspond to meaningful linguistic units. Instead, each token is represented by an atomic identifier whose internal structure and relationships with other tokens remain implicit.

This limitation is particularly relevant to Chinese and Vietnamese. Despite their different writing systems, both are tonal languages in which the syllable is a prominent unit and can be analyzed into an onset, a rime, and a lexical tone. Chinese characters do not transparently encode pronunciation \cite{coulmas1991writing}, but Pinyin provides an intermediate phonological notation. Vietnamese uses an alphabetic orthography with explicit orthographic-syllable boundaries and relatively systematic grapheme--phoneme correspondences. Character tokenization preserves Chinese syllable boundaries but treats characters as unrelated atomic entries, whereas statistical subword methods may merge syllables or fragment a Vietnamese syllable into pieces without independent linguistic interpretations. Neither paradigm explicitly represents the syllable's internal organization.

Sub-character tokenization partially addresses this issue by converting Chinese characters into Pinyin or Wubi strings before applying statistical segmentation \cite{si2023subchar}. Its final units, however, remain corpus-dependent and need not preserve syllable structure. More generally, atomic token identifiers offer little explanation of which linguistic properties are shared across input units. Explicit linguistic representations do not make neural models fully interpretable, but they provide an analyzable account of how surface text is transformed into model inputs.

We introduce \textbf{Phonemic Tokenizer}, a computational representation framework that preserves one contextual position per syllable while factorizing it into three phonological components: \textbf{onset, rime, and tone}. Each orthographic unit is first mapped to IPA and then deterministically converted into a phonological tuple. Syllables sharing a component reuse its representation, rather than being assigned entirely independent token identifiers. The framework requires neither corpus-dependent vocabulary learning nor statistical segmentation and retains syllable-level sequence length.

We instantiate the framework for Mandarin Chinese and Vietnamese. Chinese characters are assigned phrase-aware Pinyin pronunciations before IPA conversion, while Vietnamese orthographic syllables are converted directly using deterministic rules. The two instantiations share the same factorization scheme but use language-specific phonological inventories. Missing components are represented by \texttt{[EMPTY]}, and unsupported inputs are handled through character-level fallback. Including fallback and special symbols, the resulting vocabularies contain only 112 entries for Chinese and 256 for Vietnamese.

We evaluate the framework intrinsically through vocabulary size, fertility, R\'enyi efficiency, and tokenized sequence length. We then develop \textbf{PhonemicBERT} as an experimental instantiation, combining projected component embeddings with tuple-level masked prediction. For Chinese, we follow the corpus, model configurations, and evaluation protocol of \citet{si2023subchar}, enabling controlled comparisons with character, subword, and SubChar tokenization. We also evaluate robustness to synthetic homophone substitutions. For Vietnamese, we compare PhonemicBERT-Vi with established base-sized monolingual and multilingual PLMs across diverse downstream tasks. The intrinsic results demonstrate compact and efficient vocabulary utilization, while the downstream experiments show that phonemic factorization remains competitive with and outperforms established alternatives on multiple tasks across both writing systems.

Our contributions are threefold:
\begin{itemize}
    \item We propose an IPA-based computational representation that factorizes each syllable into onset, rime, and tone while preserving one contextual position and avoiding corpus-dependent segmentation.

    \item We instantiate and intrinsically evaluate the framework for Mandarin Chinese and Vietnamese, obtaining compact vocabularies across logographic and alphabetic writing systems.

    \item We introduce PhonemicBERT to evaluate the representation in pretraining, providing controlled Chinese comparisons, broad Vietnamese evaluation, and an analysis of robustness to homophone substitutions.
\end{itemize}

\section{Related Work}
\label{sec:related_work}

\subsection{Statistical and Language-Specific Tokenization}

Most pretrained language models use corpus-induced subword tokenizers such as BPE \cite{sennrich2016neural}, WordPiece \cite{schuster2012japanese}, or the Unigram Language Model \cite{kudo2018subword}. These methods balance character- and word-level vocabularies by extracting frequent patterns, but their units are not guaranteed to align with linguistic boundaries \cite{languages-bpe}. For Chinese and Vietnamese, they may merge multiple syllables or fragment a Vietnamese syllable into pieces without independent linguistic interpretations. Character tokenization avoids fragmentation in Chinese but represents phonologically related characters as unrelated atomic entries.

Vietnamese PLMs illustrate these alternatives. PhoBERT \cite{nguyen2020phobert} applies word segmentation followed by BPE, whereas WikiBERT \cite{pyysalo-etal-2021-wikibert}, mBERT \cite{devlin2019bert}, and XLM-R \cite{conneau2020unsupervised} use monolingual or multilingual statistical vocabularies. DistilBERT \cite{sanh2019distilbert} retains its teacher's tokenizer while reducing model size. None explicitly represents the onset, rime, and tone of each Vietnamese syllable. In contrast, Phonemic Tokenizer preserves one contextual position per syllable while exposing reusable phonological components.

\subsection{Linguistically Informed and Phonological Representations}

Prior work incorporates sub-character information into Chinese language models. ChineseBERT \cite{sun2021chinesebert} augments character embeddings with glyph and Pinyin features. SubChar \cite{si2023subchar} converts characters into Wubi or Pinyin strings before applying statistical subword segmentation. These approaches demonstrate the value of sub-character information, but either retain characters as the primary units or derive their final boundaries through corpus-dependent segmentation. Related work on spelling correction and robust language modeling similarly uses Pinyin as an auxiliary signal rather than as the primary structured input representation.

IPA-based tokenization has also been explored across languages. \citet{miletic-etal-2026-phonemes} convert orthographic corpora into IPA and train BPE or Unigram tokenizers over the resulting phoneme sequences, reducing script-dependent disparities and increasing cross-lingual symbol overlap. Our objective differs: we apply no statistical segmentation after IPA conversion. Instead, each syllable is deterministically represented as one onset--rime--tone tuple. Chinese and Vietnamese use separate phonological inventories; IPA provides a consistent notation for applying the same computational factorization across their different writing systems.

\subsection{Intrinsic Tokenizer Evaluation}

Downstream results jointly reflect tokenization, model architecture, pretraining data, and optimization. Intrinsic metrics therefore provide complementary information about tokenizer behavior. Fertility measures the number of tokens required to represent an input unit and quantifies fragmentation \cite{rust-etal-2021-good}, while average sequence length reflects the number of contextual positions processed by the model. R\'enyi efficiency measures how effectively the available vocabulary is utilized \cite{zouhar-etal-2023-tokenization}. However, it is not a reliable standalone predictor of downstream performance \cite{cognetta-etal-2024-two}.

We consequently evaluate Phonemic Tokenizer using vocabulary size, fertility, R\'enyi efficiency, and average sequence length, and report these properties separately from the downstream performance of PhonemicBERT.

\section{Phonemic Tokenization}
\label{sec:phonemic_tokenization}

We propose \textbf{Phonemic Tokenizer}, a computational representation framework that maps each phonologically analyzable syllable to a structured onset--rime--tone tuple. Given an input sequence
\begin{equation}
    C=[c_1,\ldots,c_n],
\end{equation}
the tokenizer produces
\begin{equation}
    P=[p_1,\ldots,p_n],
    \qquad
    p_i=(O_i,R_i,T_i),
    \label{eq:phonemic_tokenizer_output}
\end{equation}
where $O_i$, $R_i$, and $T_i$ denote the onset, rime, and tone of the syllable corresponding to $c_i$. The three components are aligned at one contextual position rather than emitted as three sequential tokens. The framework therefore exposes syllable-internal structure without tripling sequence length.

\subsection{Orthography-to-IPA Conversion}
\label{sec:ipa_conversion}

The tokenizer first converts each native input unit into a normalized IPA syllable:
\begin{equation}
    \phi_{\ell}:c_i\longmapsto s_i,
    \qquad
    \ell\in\{\mathrm{Zh},\mathrm{Vi}\},
    \label{eq:ipa_conversion}
\end{equation}
where $\ell$ denotes the language. IPA provides a consistent intermediate notation, but each language retains its own phonological inventory and conversion rules.

Vietnamese orthographic syllables are converted directly using deterministic grapheme-to-phoneme correspondences. Because pronunciation varies across regions, we adopt a composite standard based on established descriptions of Vietnamese phonology \cite{haophonetic,thuat2016}. It provides a normalized computational representation rather than a narrow transcription of a single dialect.

Chinese characters do not transparently encode pronunciation. We therefore first assign an aligned Pinyin sequence
\begin{equation}
    \Psi_{\mathrm{Zh}}:
    [c_1,\ldots,c_n]
    \longmapsto
    [q_1,\ldots,q_n],
\end{equation}
using phrase-level pronunciation lookup to resolve covered polyphonic characters. Phrase entries are matched longest-first; unmatched characters receive their default isolated pronunciation. Each selected Pinyin syllable is then deterministically converted into IPA:
\begin{equation}
    \rho_{\mathrm{Zh}}:q_i\longmapsto s_i.
\end{equation}
Pinyin thus serves only as a pronunciation-bearing bridge between Chinese orthography and IPA. Polyphonic readings covered by the phrase lexicon are resolved before factorization, whereas distinct homophonous characters may still map to the same phonemic representation. We analyze this trade-off in Section~\ref{sec:homophone_analysis}.

\subsection{Syllable Factorization}
\label{sec:syllable_factorization}

Each IPA syllable is factorized using a language-specific function
\begin{equation}
    \gamma_{\ell}:s_i\longmapsto
    (O_i,R_i,T_i).
    \label{eq:syllable_factorization}
\end{equation}
The onset contains the consonantal material preceding the rime. Internally, the tokenizer identifies the rime as
\begin{equation}
    R_i=G_i\Vert V_i\Vert F_i,
    \label{eq:rime_composition}
\end{equation}
where $G_i$, $V_i$, and $F_i$ denote an optional glide, the vowel nucleus, and an optional final. These elements are used only during parsing; their combination is retained as one atomic rime component.

The tone is represented independently. Syllables differing only in tone therefore share their onset and rime entries, while syllables sharing an onset or rime reuse the corresponding component. This explicit sharing distinguishes phonemic factorization from assigning an unrelated identifier to every character or complete syllable.

If an onset or an overt tone is absent under the language-specific convention, its slot is assigned \texttt{[EMPTY]}. This symbol is distinct from \texttt{[UNK]}, which denotes unsupported input, and \texttt{[PAD]}, which is used for sequence padding. A valid native syllable must contain a recognized rime.

\subsection{Coverage and Complexity}
\label{sec:coverage_fallback}

Units outside the native syllable inventory, including foreign strings, URLs, punctuation, and malformed text, are handled through character-level fallback. A fallback character $x$ is represented as
\begin{equation}
    p(x)=(x,x,x),
\end{equation}
preserving the three-slot interface. Characters absent from both the native and fallback inventories are mapped to \texttt{[UNK]}. Consequently, native syllables occupy one contextual position, while an unsupported multi-character unit may expand into several fallback positions.

The conversion and factorization rules are fixed independently of the pretraining corpus. With dictionary lookup and bounded phrase matching, the tokenizer runs in linear time with respect to the input length. Including fallback and special symbols, the Chinese and Vietnamese instantiations contain 112 and 256 vocabulary entries, respectively. Complete algorithms, pronunciation resources, and IPA mapping tables are provided in Appendices~\ref{sec:app_vi} and~\ref{sec:app_zh}.

\section{PhonemicBERT}
\label{sec:phonemicbert}

To evaluate the proposed representation in language model pretraining, we develop \textbf{PhonemicBERT}, a BERT encoder adapted to process onset--rime--tone tuples. The Transformer architecture remains unchanged; only the input representation and masked prediction objective are modified. We independently pretrain \textbf{PhonemicBERT-Zh} and \textbf{PhonemicBERT-Vi} using their language-specific tokenizers and vocabularies.

\subsection{Factorized Input Representation}
\label{sec:factorized_input}

Given a sequence
\begin{equation}
    P=[p_1,\ldots,p_n],
    \qquad p_i=(O_i,R_i,T_i),
\end{equation}
all three components of $p_i$ remain aligned at one contextual position. Let $\mathcal{V}_{\ell}$ be the complete vocabulary for language $\ell$, including phonological components, fallback characters, and special symbols. PhonemicBERT uses a shared embedding table
\begin{equation}
    E_{\ell}:\mathcal{V}_{\ell}\rightarrow\mathbb{R}^{d},
\end{equation}
where $d$ is the Transformer hidden size. The three component embeddings are concatenated and projected from $3d$ to $d$:
\begin{equation}
    \tilde{e}_i
    =
    W_{\mathrm{proj}}
    \left[
        E_{\ell}(O_i)
        \oplus
        E_{\ell}(R_i)
        \oplus
        E_{\ell}(T_i)
    \right]
    +b_{\mathrm{proj}},
    \label{eq:joint_phonological_projection}
\end{equation}
where $W_{\mathrm{proj}}\in\mathbb{R}^{d\times3d}$. Standard positional and token-type embeddings are then added before the sequence is passed to the BERT encoder. Thus, a sequence of $n$ syllables contains $n$, rather than $3n$, Transformer positions.

The shared table enables syllables with common components to reuse the corresponding embeddings, while the projection learns interactions among onset, rime, and tone. Fallback and special units follow the same interface: a fallback character $a$ is represented as $(a,a,a)$, while absent and unknown components use \texttt{[EMPTY]} and \texttt{[UNK]}, respectively.

\subsection{Holistic Masked Language Modeling}
\label{sec:holistic_mlm}

Masking only one component would reveal the remaining phonological information at the same position. We therefore introduce \textbf{Holistic Masked Language Modeling} (H-MLM), which masks and reconstructs the complete tuple.

We sample a set $\mathcal{M}$ containing 15\% of the sequence positions. For every $i\in\mathcal{M}$, all components are replaced simultaneously:
\begin{equation}
    (O_i,R_i,T_i)
    \longmapsto
    \left(
        \texttt{[MASK]},
        \texttt{[MASK]},
        \texttt{[MASK]}
    \right).
    \label{eq:tuple_masking}
\end{equation}
H-MLM does not use the standard 80/10/10 replacement strategy: every selected tuple is fully masked, forcing reconstruction from surrounding context.

Let $h_i\in\mathbb{R}^{d}$ be the final contextual representation at a masked position. Three independent heads predict the original components:
\begin{equation}
\begin{aligned}
    \hat{y}_i^{O} &= \operatorname{Softmax}(W_Oh_i+b_O),\\
    \hat{y}_i^{R} &= \operatorname{Softmax}(W_Rh_i+b_R),\\
    \hat{y}_i^{T} &= \operatorname{Softmax}(W_Th_i+b_T).
\end{aligned}
\label{eq:hmlm_predictions}
\end{equation}
Each head predicts over the complete vocabulary $\mathcal{V}_{\ell}$, allowing the same objective to handle native syllables and fallback characters. The training loss is the unweighted sum of the three cross-entropy losses:
\begin{equation}
    \mathcal{L}_{\mathrm{H\text{-}MLM}}
    =
    \mathcal{L}_{O}
    +
    \mathcal{L}_{R}
    +
    \mathcal{L}_{T}.
    \label{eq:hmlm_total_loss}
\end{equation}
H-MLM therefore preserves one prediction position per syllable while requiring the shared contextual representation to reconstruct its complete phonological structure.

\section{Experiments} \label{sec:experiments}

We evaluate Phonemic Tokenizer intrinsically and use PhonemicBERT to assess whether its representations support language model pretraining. Chinese provides a controlled comparison with alternative tokenization pipelines, whereas Vietnamese evaluates practical competitiveness against existing monolingual and multilingual PLMs.

\subsection{Pretraining Data and Configurations}
\label{sec:pretraining_setup}

PhonemicBERT-Zh uses the 2.3GB Baidu Baike corpus released by \citet{si2023subchar}. After separating punctuation, we retain lines in which at least 70\% of the remaining characters are Chinese, yielding 2.1GB of pretraining text. PhonemicBERT-Vi uses the 20GB corpus employed by PhoBERT \cite{nguyen2020phobert}. A Vietnamese unit is considered native when it can be factorized into a valid onset--rime--tone tuple. Lines containing at least 70\% native units after punctuation separation are retained, resulting in 18.54GB. Filtering is applied at the line level; unsupported material within retained lines is handled through character fallback.

\begin{table}[t]
    \centering
    \small
    \setlength{\tabcolsep}{4pt}
    \begin{tabular}{lcc}
    \toprule
    \textbf{Configuration} & \textbf{Zh} & \textbf{Vi} \\
    \midrule
    Source size              & 2.3 GB & 20 GB \\
    Retained size            & 2.1 GB & 18.54 GB \\
    Transformer layers       & 6 / 12 & 12 \\
    Hidden size              & 768 & 768 \\
    Attention heads          & 12 & 12 \\
    FFN size                 & 3,072 & 3,072 \\
    Vocabulary size          & 112 & 256 \\
    Maximum length           & 512 & 512 \\
    Batch size               & 256 & 256 \\
    Learning rate            & $5\times10^{-4}$ & $5\times10^{-4}$ \\
    Training steps           & 1M & 1M \\
    \bottomrule
    \end{tabular}
    \caption{Pretraining data and model configurations. Both models use AdamW, 1\% warmup, and 15\% tuple masking.}
    \label{tab:pretraining_configuration}
\end{table}

For Chinese, we follow the model capacities, training schedule, and evaluation protocol of \citet{si2023subchar}, using six- and twelve-layer encoders. PhonemicBERT-Vi uses a twelve-layer BERT-base encoder. Complete preprocessing and optimization details are provided in Table \ref{tab:pretraining_configuration}.

\subsection{Intrinsic Tokenizer Evaluation}
\label{sec:intrinsic_tokenizer_evaluation}

For Chinese, we compare Phonemic Tokenizer with Character, Subword, SubChar-Wubi, and SubChar-Pinyin tokenizers from \citet{si2023subchar}. For Vietnamese, we compare it with the released tokenizers of PhoBERT \cite{nguyen2020phobert}, WikiBERT \cite{pyysalo-etal-2021-wikibert}, mBERT \cite{devlin2019bert}, and XLM-R \cite{conneau2020unsupervised}. Model-specific preprocessing, including PhoBERT word segmentation, is retained, while model control symbols are excluded from all measurements.

We report vocabulary size, R\'enyi efficiency, and average tokenized sequence length for both languages. For Vietnamese, we additionally report fertility:
\begin{equation}
    \operatorname{Fertility}(\tau)
    =
    \frac{\sum_{x\in\mathcal{D}}N_{\mathrm{pos}}(\tau(x))}
         {\sum_{x\in\mathcal{D}}N_{\mathrm{unit}}(x)},
    \label{eq:fertility}
\end{equation}
where $N_{\mathrm{unit}}(x)$ counts whitespace-delimited input units and $N_{\mathrm{pos}}$ counts output contextual positions. One phonemic tuple counts as one position, while fallback may expand a non-native unit into multiple character positions.

Following \citet{zouhar-etal-2023-tokenization}, R\'enyi efficiency is computed as
\begin{equation}
    \operatorname{RE}_{\alpha}(\tau)
    =
    \frac{
        \frac{1}{1-\alpha}
        \log\sum_{v\in\mathcal{V}_{\tau}}p(v)^{\alpha}
    }{
        \log|\mathcal{V}_{\tau}|
    },
    \qquad \alpha=2.5,
    \label{eq:renyi_efficiency}
\end{equation}
where $p(v)$ is the empirical unigram probability of vocabulary entry $v$. For Phonemic Tokenizer, component occurrences in all three aligned slots are included in this distribution, although their tuple contributes only one contextual position to sequence length. We treat R\'enyi efficiency as a descriptive intrinsic property rather than a standalone predictor of downstream performance \cite{cognetta-etal-2024-two}.

\subsection{Pretrained Baselines}
\label{sec:baselines}

For Chinese, we compare PhonemicBERT-Zh with the CharTokenizer, Subword, SubChar-Wubi, and SubChar-Pinyin models reported by \citet{si2023subchar}. These represent atomic character input, character-level BPE, glyph-based SubChar tokenization, and Pinyin-based SubChar tokenization, respectively. The comparison matches the source corpus, encoder capacities, training schedule, and downstream protocol as closely as possible. Because PhonemicBERT also introduces factorized projection and H-MLM, the results compare complete tokenization-and-pretraining pipelines rather than changing only a scalar tokenizer.

For Vietnamese, we compare PhonemicBERT-Vi with WikiBERT, PhoBERT, mBERT, DistilBERT \cite{sanh2019distilbert}, and XLM-R-base. These models differ in pretraining data and objectives; the comparison therefore evaluates practical PLM performance rather than providing a controlled tokenizer ablation. Baseline results are taken from VLUE \cite{do-etal-2024-vlue}, except PhoNER and VietMed results, which follow \citet{dinh-etal-2026-morphology}. We exclude CafeBERT and HuTieuBERT because they differ substantially in model scale, continued-pretraining history, or encoder architecture. 

\subsection{Downstream Benchmarks}
\label{sec:downstream_benchmarks}

The Chinese evaluation covers eleven benchmarks: TNEWS, IFLYTEK, and THUCNews for text classification; AFQMC, BQ, and CSL for sentence-pair understanding; OCNLI and CLUEWSC for inference and coreference; C3 and CMRC for reading comprehension; and CLUENER for named-entity recognition. Following baseline availability, the twelve-layer comparison covers the nine benchmarks reported by \citet{si2023subchar}, while CMRC and CLUENER are evaluated only with the six-layer configuration.

The Vietnamese evaluation covers twelve datasets: VSMEC, ViHSD, ViHOS, UIT-ViSFD, ViOCD, ViCTSD, and UIT-VSFC for social-media classification and span detection; NIIVTB POS for part-of-speech tagging; PhoNER and VietMed for general- and medical-domain named-entity recognition; ViNLI for natural language inference; and ViNewsQA for extractive question answering. 

\subsection{Fine-Tuning and Reporting}
\label{sec:finetuning_evaluation}

For Chinese, we follow \citet{si2023subchar}: the official development set is used as the fixed test set, while each run samples 10\% of the official training set as a new development set. We report accuracy for sequence-level tasks, exact match for CMRC, and entity-level F1 for CLUENER.

For Vietnamese, we use the official splits and fine-tuning setup of VLUE \cite{do-etal-2024-vlue}. Classification, NLI, and NIIVTB POS use macro-F1; ViHOS uses token-level macro-F1; PhoNER and VietMed use span-level F1; and ViNewsQA uses extractive QA F1.

We fine-tune PhonemicBERT with four random seeds for every task and report the mean and standard deviation. For Chinese, each seed also produces a new random development split. Published baseline are reported by their source studies and may not use the same number of runs.

\section{Intrinsic Evaluation of the Phonemic Tokenizer} \label{sec:intrinsic-evaluation}

\subsection{Vietnamese Tokenization Results} \label{sec:intrinsic_results_vi}

We compare the Vietnamese Phonemic Tokenizer with the released tokenizers of PhoBERT, WikiBERT, mBERT, and XLM-RoBERTa. The evaluation considers vocabulary compactness, syllable-normalized Fertility, Rényi efficiency, and average tokenized sequence length. We subsequently analyze the occurrence of units that cannot be phonologically factorized as standard Vietnamese syllables.

\subsubsection{Vocabulary compactness}

\begin{table}[ht]
    \centering
    \small
    \begin{tabular}{lrr}
    \toprule
    \textbf{Tokenizer} & \textbf{Vocabulary} & \textbf{Pretraining corpus} \\
    \midrule
    PhoBERT       & 64,000  & 20 GB \\
    WikiBERT      & 30,000  & - \\
    mBERT         & 119,537 & - \\
    XLM-RoBERTa   & 250,002 & 2.5 TB \\
    Phonemic      & \textbf{256} & 18.54 GB \\
    \bottomrule
    \end{tabular}
    \caption{Vocabulary sizes and pretraining-corpus scales of the Vietnamese tokenizers. Corpus sizes describe the complete pretraining resources of the corresponding models and are not directly comparable across monolingual and multilingual settings.}
    \label{tab:vi_tokenizer_overview}
\end{table}

Table~\ref{tab:vi_tokenizer_overview} compares the vocabulary sizes and pretraining-corpus scales of the evaluated tokenizers. The Phonemic Tokenizer uses only 256 vocabulary entries, including phonological components, fallback characters, and special symbols. This vocabulary is approximately 117$\times$ smaller than that of WikiBERT, 250$\times$ smaller than that of PhoBERT, 467$\times$ smaller than that of mBERT, and 977$\times$ smaller than that of XLM-RoBERTa.

The compact vocabulary results from representing a syllable compositionally through shared onset, rime, and tone entries rather than assigning an atomic entry to every syllable or statistically induced subword. Corpus scale is reported only to contextualize the pretrained tokenizers and should not be interpreted as a controlled comparison: multilingual corpora contain different proportions of Vietnamese data, and their total sizes are not directly comparable with the Vietnamese-only corpus used by PhonemicBERT-Vi.

\subsubsection{Syllable-level Fertility}

\begin{table*}[htp]
    \centering
    \resizebox{\linewidth}{!}{
    \begin{tabular}{clccccc}
    \toprule
    \textbf{\#} & \textbf{Dataset} & \textbf{PhoBERT} & \textbf{WikiBERT} & \textbf{mBERT} & \textbf{XLM-R} & \textbf{Phonemic} \\
    \midrule
    1  & VSMEC & 1.0882 & 1.3356 & 1.2771 & 1.1848 & \textbf{1.0562} \\
    2  & ViHOS & 2.4561 & 1.2973 & 1.4405 & 1.3880 & \textbf{1.0116} \\
    3  & ViHSD & 2.4374 & 1.3254 & 1.4394 & 1.3942 & \textbf{1.0280} \\
    4  & NIIVTB POS & \textbf{1.0200} & 1.3426 & 1.1694 & 1.1674 & 1.0422 \\
    5  & PhoNER & \textbf{1.0341} & 1.3611 & 1.1558 & 1.1834 & 1.0974 \\
    6  & UIT-ViSFD & 1.2428 & 1.5039 & 1.3991 & 1.2948 & \textbf{1.1643} \\
    7  & ViOCD & 2.7261 & 1.3320 & 1.4502 & 1.3747 & \textbf{1.0139} \\
    8  & ViCTSD & 1.2122 & 1.4733 & 1.3442 & 1.2312 & \textbf{1.1388} \\
    9  & UIT-VSFC & 1.0208 & 1.3571 & 1.0842 & 1.1381 & \textbf{1.0062} \\
    10 & VietMed & 1.0142 & 1.3082 & 1.1114 & 1.0413 & \textbf{1.0007} \\
    11 & ViNewsQA & 1.1930 & 1.5058 & 1.3109 & 1.2004 & \textbf{1.1363} \\
    12 & ViNLI & 1.2286 & 1.5314 & 1.3060 & 1.2292 & \textbf{1.1678} \\
    \midrule
    13 & Standard Dictionary & 1.0660 & 1.4019 & 1.4887 & 1.3798 & \textbf{1.0000} \\
    \bottomrule
    \end{tabular}}
    \caption{Syllable-normalized Fertility on Vietnamese downstream datasets and the Standard Dictionary. Lower values indicate fewer contextual positions per input syllabic unit. For the Phonemic Tokenizer, an onset--rime--tone tuple counts as one contextual position. Values below one for statistical tokenizers can result from merging multiple syllabic units into a single token.}
    \label{tab:vi_Fertility}
\end{table*}

Table~\ref{tab:vi_Fertility} reports the average number of contextual positions produced per whitespace-delimited input unit. The Phonemic Tokenizer obtains the lowest Fertility on 10 of the 12 downstream datasets. The only exceptions are NIIVTB POS and PhoNER, on which PhoBERT achieves slightly lower values: 1.0200 versus 1.0422 on NIIVTB POS and 1.0341 versus 1.0974 on PhoNER. Across the remaining datasets, the Phonemic Tokenizer maintains Fertility between 1.0007 and 1.1678, indicating that most input units are represented using one contextual position.

The advantage is particularly pronounced on ViHOS, ViHSD, and ViOCD. PhoBERT produces 2.4561, 2.4374, and 2.7261 positions per input unit on these datasets, respectively, whereas the Phonemic Tokenizer produces only 1.0116, 1.0280, and 1.0139. WikiBERT, mBERT, and XLM-RoBERTa also exhibit greater fragmentation on most datasets, with Fertility commonly ranging from approximately 1.2 to 1.5. These results suggest that the statistical vocabularies of the pretrained tokenizers are more sensitive to variation across domains and writing styles, whereas deterministic syllable factorization provides more stable positional behavior.

The Fertility of the Phonemic Tokenizer nevertheless exceeds one on natural text because not every whitespace-delimited unit is a well-formed Vietnamese syllable. Social-media datasets such as VSMEC, ViHOS, ViHSD, UIT-ViSFD, ViOCD, ViCTSD, and UIT-VSFC contain teen-code spellings, non-standard abbreviations, expressive character sequences, and emoticons. Such forms cannot always be factorized into valid onset--rime--tone tuples and are therefore represented through character-level fallback. The degree of expansion varies across these datasets because it depends on both the frequency and character length of the unsupported units. For example, UIT-ViSFD and ViCTSD yield Fertility values of 1.1643 and 1.1388, whereas ViHOS, ViHSD, and ViOCD remain much closer to one.

Elevated Fertility is also observed on PhoNER, ViNewsQA, and ViNLI, with values of 1.0974, 1.1363, and 1.1678, respectively. These corpora contain online-news or Wikipedia-derived text and consequently include foreign names, transliterated entities, technical terminology, and other non-Vietnamese expressions. Character-level fallback preserves these units without forcing them into the Vietnamese phonological inventory, but may expand one input unit into multiple contextual positions.

\subsubsection{Vocabulary utilization}

\begin{table*}[ht]
    \centering
    \resizebox{\linewidth}{!}{
    \begin{tabular}{clccccc}
    \toprule
    \textbf{\#} & \textbf{Dataset} & \textbf{PhoBERT} & \textbf{WikiBERT} & \textbf{mBERT} & \textbf{XLM-R} & \textbf{Phonemic} \\
    \midrule
    1  & VSMEC       & 0.4740 & 0.4876 & 0.4354 & 0.3863 & \textbf{0.5661} \\
    2  & ViHOS       & 0.2702 & 0.4923 & 0.1701 & 0.3535 & \textbf{0.5777} \\
    3  & ViHSD       & 0.2710 & 0.4945 & 0.1720 & 0.3572 & \textbf{0.5819} \\
    4  & NIIVTB POS  & 0.4405 & 0.4706 & 0.4150 & 0.3412 & \textbf{0.5300} \\
    5  & PhoNER      & 0.3831 & 0.4315 & 0.3734 & 0.3046 & \textbf{0.4692} \\
    6  & UIT-ViSFD   & 0.4895 & 0.4481 & 0.4006 & 0.3907 & \textbf{0.4996} \\
    7  & ViOCD       & 0.2596 & 0.4570 & 0.1471 & 0.3301 & \textbf{0.5257} \\
    8  & ViCTSD      & 0.5020 & 0.4743 & 0.4282 & 0.4140 & \textbf{0.5410} \\
    9  & UIT-VSFC    & 0.3701 & 0.4225 & 0.3572 & 0.2885 & \textbf{0.4643} \\
    10 & VietMed     & 0.4331 & 0.4672 & 0.4211 & 0.3890 & \textbf{0.5440} \\
    11 & ViNewsQA    & 0.4529 & 0.4699 & 0.4235 & 0.3985 & \textbf{0.5403} \\
    12 & ViNLI       & 0.4590 & 0.4772 & 0.4158 & 0.3981 & \textbf{0.5319} \\
    \midrule
    13 & Standard Dictionary
                      & 0.6420 & 0.4874 & 0.4614 & 0.4461 & \textbf{0.8232} \\
    \bottomrule
    \end{tabular}}
    \caption{Rényi efficiency of Vietnamese tokenizers. Higher values indicate more balanced utilization of the corresponding vocabulary.}
    \label{tab:vi_renyi}
\end{table*}

Table~\ref{tab:vi_renyi} reports Rényi efficiency. The Phonemic Tokenizer achieves the highest value on every evaluated dataset and on the Standard Dictionary. Its advantage is particularly pronounced on ViHOS, ViHSD, ViOCD, and the dictionary, where the distributions induced by several pretrained tokenizers are comparatively concentrated.

These results indicate that the compact phonemic inventory is used more evenly than the substantially larger statistical vocabularies. The dictionary result of 0.8232 further shows that the component inventory provides broad and balanced coverage of well-formed Vietnamese syllables. Nevertheless, Rényi efficiency measures vocabulary-distribution properties rather than downstream quality and should be interpreted jointly with Fertility, sequence length, and the PhonemicBERT results.

\subsubsection{Average tokenized sequence length}

\begin{table}[htp]
    \centering
    \resizebox{\linewidth}{!}{
    \begin{tabular}{clcccccc}
    \toprule
    \textbf{\#} & \textbf{Dataset} & \textbf{Input} & \textbf{PhoBERT} & \textbf{WikiBERT} & \textbf{mBERT} & \textbf{XLM-R} & \textbf{Phonemic} \\
    \midrule
    1 & VSMEC & 14.01 & 15.24 (+01.23) & 18.71 (+04.70) & 17.89 (+03.88) & 16.59 (+02.58) & \textbf{14.79 (+00.78)} \\
    2 & ViHOS & 14.32 & 35.16 (+20.84) & 18.57 (+04.25) & 20.62 (+06.30) & 19.87 (+05.55) & \textbf{14.48 (+00.16)} \\
    3 & ViHSD & 13.11 & 31.96 (+18.85) & 17.38 (+04.27) & 18.87 (+05.76) & 18.28 (+05.17) & \textbf{13.48 (+00.37)} \\
    4 & NIIVTB POS & 28.86 & \textbf{29.43 (+00.57)} & 38.74 (+09.88) & 33.74 (+04.88) & 33.68 (+04.82) & 30.07 (+01.21) \\
    5 & PhoNER & 34.63 & \textbf{35.81 (+01.18)} & 47.13 (+12.50) & 40.02 (+05.39) & 40.98 (+06.35) & 38.00 (+03.37) \\
    6 & UIT-ViSFD & 36.19 & 44.97 (+08.78) & 54.42 (+18.23) & 50.63 (+14.44) & 46.85 (+10.66) & \textbf{42.13 (+05.94)} \\
    7 & ViOCD & 31.77 & 86.61 (+54.84) & 42.32 (+10.55) & 46.07 (+14.30) & 43.67 (+11.90) & \textbf{32.21 (+00.44)} \\
    8 & ViCTSD & 29.38 & 35.62 (+06.24) & 43.29 (+13.91) & 39.50 (+10.12) & 36.18 (+06.80) & \textbf{33.46 (+04.08)} \\
    9 & UIT-VSFC & 14.23 & 14.52 (+00.29) & 19.31 (+05.08) & 15.43 (+01.20) & 16.19 (+01.96) & \textbf{14.32 (+00.09)} \\
    10 & VietMed & 23.37 & 23.71 (+00.34) & 30.58 (+07.21) & 25.98 (+02.61) & 24.34 (+00.97) & \textbf{23.39 (+00.02)} \\
    11 & ViNewsQA & 39.12 & 46.67 (+07.55) & 58.90 (+19.78) & 51.28 (+12.16) & 46.96 (+07.84) & \textbf{44.45 (+05.33)} \\
    12 & ViNLI & 25.05 & 30.77 (+05.72) & 38.36 (+13.31) & 32.71 (+07.66) & 30.79 (+05.74) & \textbf{29.25 (+04.20)} \\
    \bottomrule
    \end{tabular}}
    \caption{Average input and tokenized sequence lengths on Vietnamese downstream datasets. Parenthesized values indicate the absolute change relative to the average input length. A positive value denotes sequence expansion. Bold indicates the shortest tokenized sequence for each dataset. Lengths count contextual positions and exclude model-specific special tokens.}
    \label{tab:vi_average_length}
\end{table}

Table~\ref{tab:vi_average_length} reports the average number of contextual positions per example. The results closely follow the Fertility patterns. The Phonemic Tokenizer produces the shortest average sequences on 10 of the 12 datasets, matching the datasets on which it obtains the lowest Fertility. PhoBERT produces slightly shorter sequences only on NIIVTB POS and PhoNER.

Larger increases occur on datasets containing more non-standard or non-Vietnamese material. The Phonemic Tokenizer expands UIT-ViSFD by 5.94 positions, ViNewsQA by 5.33, ViNLI by 4.20, and PhoNER by 3.37 on average. These increases are consistent with the use of character-level fallback for foreign terminology, named entities, URLs, and malformed or non-standard forms. Nevertheless, the resulting sequences remain shorter than those produced by the other evaluated tokenizers on UIT-ViSFD, ViNewsQA, and ViNLI.

\subsubsection{Type-level coverage of native Vietnamese units}

\begin{table}[t]
    \centering
    \resizebox{0.6\textwidth}{!}{
    \begin{tabular}{lrrr}
    \toprule
    \textbf{Dataset} & \textbf{Unique types} & \textbf{Vietnamese} & \textbf{Coverage} \\
    \midrule
    VSMEC      & 4,750  & 3,577 & 75\% \\
    ViHOS      & 7,301  & 4,455 & 61\% \\
    ViHSD      & 11,852 & 5,624 & 47\% \\
    NIIVTB POS & 8,316  & 4,698 & 56\% \\
    PhoNER     & 2,706  & 2,122 & 78\% \\
    UIT-ViSFD  & 6,473  & 3,775 & 58\% \\
    ViOCD      & 4,598  & 3,068 & 67\% \\
    ViCTSD     & 6,935  & 4,289 & 62\% \\
    UIT-VSFC   & 2,451  & 2,019 & 82\% \\
    VietMed    & 2,763  & 2,480 & \textbf{90\%} \\
    ViNewsQA   & 13,820 & 4,742 & 34\% \\
    ViNLI      & 11,298 & 4,366 & 39\% \\
    \bottomrule
    \end{tabular}}
    \caption{Type-level coverage of phonologically analyzable Vietnamese units. Coverage is the proportion of unique observed types recognized as standard Vietnamese syllables.}
    \label{tab:vi_type_coverage}
\end{table}

To understand the effect of non-Vietnamese material, we first examine the number of unique units in each dataset. Table~\ref{tab:vi_type_coverage} reports the total number of distinct observed types, the number that can be phonologically factorized as standard Vietnamese syllables, and their percentage.

Type-level coverage varies substantially across datasets. VietMed exhibits the highest coverage at 90\%, followed by UIT-VSFC at 82\% and PhoNER at 78\%. In contrast, only 34\% and 39\% of the distinct types in ViNewsQA and ViNLI, respectively, are recognized as standard Vietnamese. These low type-level percentages indicate a long tail of names, foreign expressions, numerals, symbols, malformed strings, and other dataset-specific forms that cannot be analyzed through the native syllable inventory.

\subsubsection{Token-level coverage of native Vietnamese units}

\begin{table}[t]
    \centering
    \resizebox{0.6\textwidth}{!}{
    \begin{tabular}{lrrr}
    \toprule
    \textbf{Dataset} & \textbf{Occurrences} & \textbf{Vietnamese} & \textbf{Coverage} \\
    \midrule
    VSMEC      & 102,462    & 85,080     & 83\% \\
    ViHOS      & 160,121    & 124,233    & 78\% \\
    ViHSD      & 450,182    & 333,658    & 74\% \\
    NIIVTB POS & 617,413    & 503,242    & 82\% \\
    PhoNER     & 381,038    & 264,019    & 69\% \\
    UIT-ViSFD  & 468,599    & 350,340    & 75\% \\
    ViOCD      & 176,651    & 148,720    & 84\% \\
    ViCTSD     & 334,603    & 276,280    & 83\% \\
    UIT-VSFC   & 231,582    & 200,519    & 87\% \\
    VietMed    & 216,753    & 214,818    & \textbf{99\%} \\
    ViNewsQA   & 2,754,214  & 2,321,240  & 84\% \\
    ViNLI      & 14,998,707 & 11,500,830 & 77\% \\
    \bottomrule
    \end{tabular}}
    \caption{Token-level coverage of phonologically analyzable Vietnamese units. Unlike Table~\ref{tab:vi_type_coverage}, repeated occurrences are retained.}
    \label{tab:vi_token_coverage}
\end{table}

Type counts assign the same weight to frequent and rare forms. We therefore also compute token-level coverage, counting repeated occurrences. Table~\ref{tab:vi_token_coverage} shows that native Vietnamese syllables constitute the majority of observed occurrences in every dataset. Coverage ranges from 69\% on PhoNER to 99\% on VietMed.

The difference between type- and token-level coverage is especially large on ViNewsQA and ViNLI. Although only 34\% and 39\% of their unique types are recognized as standard Vietnamese, these types account for 84\% and 77\% of all occurrences, respectively. Thus, much of the unsupported vocabulary belongs to a diverse but relatively infrequent tail, whereas frequently occurring text remains predominantly analyzable. Nevertheless, every unsupported occurrence may invoke character-level fallback, explaining why the Fertility of the Phonemic Tokenizer can exceed one even when native Vietnamese accounts for most corpus tokens.

\subsubsection{Overall findings}

The Vietnamese intrinsic evaluation reveals three complementary properties. First, phonological factorization reduces the vocabulary from tens or hundreds of thousands of entries to 256 shared components. Second, the resulting vocabulary is used more evenly, as reflected by consistently higher Rényi efficiency. Third, the tokenizer preserves one contextual position for every analyzable Vietnamese syllable, while deviations from unit Fertility arise primarily from non-native or malformed material handled through fallback. The type- and token-level analyses further show that such material is diverse but often infrequent, allowing the tokenizer to maintain broad operational coverage without incorporating it into the native phonological inventory.

\subsection{Chinese Tokenization Results} \label{sec:intrinsic_results_zh}

We compare the Chinese Phonemic Tokenizer with Character, Subword, SubChar-Wubi, and SubChar-Pinyin tokenization. Because the Chinese input and the corresponding phonemic representation are aligned at the character level, we do not separately report Fertility. For the Phonemic and Character tokenizers, each input character deterministically contributes one contextual position, while the Fertility of Subword and SubChar tokenizers would primarily restate the sequence-compression patterns already captured by average tokenized length. We therefore focus on Rényi efficiency, sequence length, vocabulary size, and native-character coverage.

\subsubsection{Rényi efficiency}

\begin{table*}[th]
    \centering
    \resizebox{\linewidth}{!}{
    \begin{tabular}{clccccc}
    \toprule
    \textbf{\#} & \textbf{Dataset} & \textbf{Character} & \textbf{Subword} & \textbf{SubChar-Wubi} & \textbf{SubChar-Pinyin} & \textbf{Phonemic} \\
    \midrule
    1  & TNEWS     & 0.5002 & 0.4920 & 0.4927 & 0.5158 & \textbf{0.6993} \\
    2  & IFLYTEK   & 0.5242 & 0.5389 & 0.5507 & 0.5497 & \textbf{0.6861} \\
    3  & CLUEWSC   & 0.4376 & 0.4688 & 0.5206 & 0.5063 & \textbf{0.6745} \\
    4  & AFQMC     & 0.3741 & 0.3667 & 0.3256 & 0.3482 & \textbf{0.6607} \\
    5  & OCNLI     & 0.4772 & 0.4915 & 0.5257 & 0.5202 & \textbf{0.6651} \\
    6  & CSL       & 0.5064 & 0.5100 & 0.5359 & 0.5096 & \textbf{0.6874} \\
    7  & C3        & 0.4867 & 0.4825 & 0.5026 & 0.4639 & \textbf{0.6713} \\
    8  & THUCNews  & 0.4873 & 0.5078 & 0.5330 & 0.5264 & \textbf{0.6961} \\
    9  & BQ        & 0.4529 & 0.4488 & 0.4668 & 0.4515 & \textbf{0.6789} \\
    10 & CMRC      & 0.4982 & 0.5264 & 0.5451 & 0.5550 & \textbf{0.7099} \\
    11 & CLUENER   & 0.4988 & 0.5149 & 0.5261 & 0.5351 & \textbf{0.6970} \\
    \bottomrule
    \end{tabular}}
    \caption{Rényi efficiency of Chinese tokenizers. Higher values indicate
    more balanced utilization of the corresponding vocabulary. For the Phonemic
    Tokenizer, the distribution is computed over identifiers occurring in the
    three aligned component slots.}
    \label{tab:zh_renyi}
\end{table*}

Table~\ref{tab:zh_renyi} reports Rényi efficiency on eleven Chinese benchmarks. The Phonemic Tokenizer achieves the highest efficiency on every dataset, with values ranging from 0.6607 on AFQMC to 0.7099 on CMRC. By contrast, the strongest value among the four baselines ranges from 0.3741 to 0.5550. The advantage is observed consistently across short-text classification, long-document classification, sentence-pair understanding, reading comprehension, and sequence-labeling datasets.

The largest difference occurs on AFQMC, where the Phonemic Tokenizer obtains 0.6607, compared with 0.3741 for the strongest baseline. Substantial differences are also observed on BQ (0.6789 versus 0.4668), TNEWS (0.6993 versus 0.5158), and C3 (0.6713 versus 0.5026). These results indicate that the compact onset--rime--tone inventory is used more evenly than the much larger statistical or character vocabularies.


\subsubsection{Average tokenized sequence length}

\begin{table*}[ht]
    \centering
    \small
    \resizebox{\linewidth}{!}{
    \begin{tabular}{clrrrrrr}
    \toprule
    \textbf{\#} & \textbf{Dataset} & \textbf{Input} & \textbf{Character} & \textbf{Subword} & \textbf{SubChar-Wubi} & \textbf{SubChar-Pinyin} & \textbf{Phonemic} \\
    \midrule
    1  & TNEWS & 22.24 & 22.24 & 20.32 & \textbf{15.96} & 16.18 & 22.24 \\
    2  & IFLYTEK & 288.17 & 288.17 & 254.35 & \textbf{182.46} & 184.49 & 288.17 \\
    3  & CLUEWSC & 66.64 & 66.64 & 61.81 & 52.27 & \textbf{51.93} & 66.64 \\
    4  & AFQMC & 13.35 & 13.35 & 12.51 & 10.98 & \textbf{10.53} & 13.35 \\
    5  & OCNLI & 17.82 & 17.82 & 15.58 & \textbf{12.03} & 12.41 & 17.82 \\
    6  & CSL & 206.08 & 206.08 & 176.40 & \textbf{125.39} & 130.61 & 206.08 \\
    7  & C3 & 6.41 & 6.41 & 6.00 & \textbf{4.45} & 4.64 & 6.41 \\
    8  & THUCNews & 851.29 & 851.29 & 741.83 & \textbf{541.62} & 552.27 & 851.29 \\
    9  & BQ & 11.86 & 11.86 & 10.85 & \textbf{8.25} & 8.54 & 11.86 \\
    10 & CMRC & 262.02 & 262.02 & 230.67 & \textbf{170.38} & 171.97 & 262.02 \\
    11 & CLUENER & 37.39 & 37.39 & 32.93 & \textbf{24.65} & 25.02 & 37.39 \\
    \bottomrule
    \end{tabular}}
    \caption{Average input and tokenized sequence lengths on Chinese benchmarks. Lengths count contextual positions and exclude model-specific special tokens. Bold indicates the shortest tokenized sequence. The Phonemic and  Character tokenizers preserve one position per input character.}
    \label{tab:zh_average_length}
\end{table*}

Table~\ref{tab:zh_average_length} compares the average number of contextual positions produced by each tokenizer. The Phonemic Tokenizer and Character tokenizer preserve the original sequence length on every dataset because each Chinese character occupies exactly one contextual position. For example, the average input length on TNEWS is 22.24 characters, and both tokenizers produce 22.24 positions. The same one-to-one correspondence holds for long inputs such as THUCNews, where all three values are 851.29.

Subword and SubChar tokenizers produce shorter sequences by merging recurring character or sub-character patterns into larger units. SubChar-Wubi produces the shortest sequences on nine datasets, while SubChar-Pinyin is shortest on CLUEWSC and AFQMC. Both substantially compress the character sequences through statistical merging. On IFLYTEK, for example, SubChar-Wubi reduces the average length from 288.17 to 182.46, whereas SubChar-Pinyin produces 184.49 positions. On THUCNews, the corresponding reductions are from 851.29 to 541.62 and 552.27 positions.

These results expose a deliberate difference in design objectives. SubChar tokenization uses part of its vocabulary capacity to store frequent multi-character combinations and thereby improve sequence compression. The Phonemic Tokenizer instead preserves character--syllable alignment while reducing vocabulary size through within-syllable factorization. It therefore does not shorten Chinese sequences, but it avoids splitting a character into multiple sequential units and maintains direct alignment for character-level tasks such as NER and span extraction.

\subsubsection{Vocabulary compactness}

\begin{table}[th]
    \centering
    \small
    \begin{tabular}{lr}
    \toprule
    \textbf{Tokenizer} & \textbf{Vocabulary size} \\
    \midrule
    Character       & 22,675 \\
    Subword         & 22,675 \\
    SubChar-Wubi    & 22,675 \\
    SubChar-Pinyin  & 22,675 \\
    Phonemic        & 112 \\
    \bottomrule
    \end{tabular}
    \caption{Vocabulary sizes of the Chinese tokenizers under the controlled
    configuration. The Phonemic count includes phonological, fallback, and
    special entries.}
    \label{tab:zh_vocabulary}
\end{table}

Table~\ref{tab:zh_vocabulary} compares the vocabulary sizes of the evaluated tokenizers. Character, Subword, SubChar-Wubi, and SubChar-Pinyin each use 22,675 entries under the controlled configuration of \citet{si2023subchar}. The Phonemic Tokenizer requires only 112 entries, including its phonological components, fallback characters, and special symbols. Its vocabulary is therefore approximately 202 times smaller, corresponding to a reduction of approximately 99.51\% relative to the baseline vocabulary size. This reduction is achieved without assigning an atomic vocabulary entry to every Chinese character. Instead, all covered characters are represented compositionally through their onset, rime, and tone. The vocabulary result and sequence-length result capture different efficiency dimensions: the Phonemic Tokenizer substantially reduces the number of embedding entries, whereas SubChar primarily reduces the number of contextual positions.

\subsubsection{Overall findings}

The Chinese intrinsic evaluation reveals a trade-off between vocabulary compactness and sequence compression. The Phonemic Tokenizer preserves the same sequence length as character tokenization while reducing the vocabulary from 22,675 to 112 entries and achieving substantially higher Rényi efficiency across all datasets. SubChar tokenizers produce shorter sequences through cross-character statistical merging, but retain a vocabulary more than 200 times larger. The proposed tokenizer therefore improves representational compactness and vocabulary utilization without sacrificing the one-to-one alignment between Chinese characters and contextual positions.

\section{Downstream Evaluation of PhonemicBERT} \label{sec:downstream_results}

\subsection{Vietnamese PLM Comparison} \label{sec:downstream_results_vi}

\subsubsection{Token- and span-level linguistic analysis}

\begin{table}[ht]
    \centering
    \resizebox{\linewidth}{!}{
    \begin{tabular}{lcccc}
    \toprule
    \textbf{Model} & \textbf{ViHOS} & \textbf{NIIVTB POS} & \textbf{VietMed} & \textbf{PhoNER} \\
    \midrule
    WikiBERT & \underline{77.05} & 75.52 & -- & -- \\
    PhoBERT & 75.69 & 77.60 & $56.04\pm4.44$ & $\underline{83.83}\pm0.15$ \\
    mBERT & 76.22 & 81.34 & $59.81\pm0.57$ & $72.82\pm0.43$ \\
    DistilBERT & 75.72 & 80.05 & -- & -- \\
    XLM-RoBERTa & 74.67 & \underline{81.76} & $\mathbf{61.76}\pm0.88$ & $69.75\pm0.38$ \\
    \midrule
    PhonemicBERT-Vi & $\mathbf{83.40}\pm1.28$ & $\mathbf{82.64}\pm1.02$ & $\underline{60.20}\pm0.84$ & $\mathbf{84.66}\pm1.10$ \\
    \bottomrule
    \end{tabular}}
    \caption{Results on Vietnamese token- and span-level tasks. ViHOS uses token-level macro-F1, NIIVTB POS uses macro-F1, and VietMed and PhoNER use span-level F1. PhonemicBERT-Vi results are means over four random seeds. Bold and underlining indicate the highest and second-highest reported means, respectively. Results on ViHOS and NIIVTB POS are reported from \cite{do-etal-2024-vlue}, while results on the VietMed and PhoNER are from \cite{dinh-etal-2026-morphology}.}
    \label{tab:vi_sequence_results}
\end{table}

Table~\ref{tab:vi_sequence_results} reports results on tasks requiring predictions aligned with individual input positions or textual spans. PhonemicBERT-Vi achieves the highest reported mean on ViHOS, NIIVTB POS, and PhoNER, while ranking second on the medical-domain VietMed dataset.

The clearest improvement occurs on ViHOS, where PhonemicBERT-Vi obtains $83.40\pm1.28$, exceeding WikiBERT, the strongest reported baseline, by 6.35 points. On NIIVTB POS, it reaches $82.64\pm1.02$, compared with 81.76 for XLM-RoBERTa. PhonemicBERT-Vi also obtains the highest reported PhoNER mean at $84.66\pm1.10$, although its 0.83-point difference from PhoBERT is small relative to the variation across runs and should be interpreted as competitive rather than conclusive superiority.

These results are consistent with the one-position-per-syllable design of the Phonemic Tokenizer. Token- and span-level annotations can be aligned directly with syllable representations without reconstructing labels from fragmented subwords. Factorization also allows syllables sharing an onset, rime, or tone to reuse component embeddings. Nevertheless, the experiments evaluate the complete PhonemicBERT pipeline and do not isolate the effects of syllable alignment, factorized projection, and H-MLM.

On VietMed, PhonemicBERT-Vi obtains $60.20\pm0.84$, ranking second behind XLM-RoBERTa at $61.76\pm0.88$ but outperforming the reported PhoBERT and mBERT means. The broader multilingual pretraining of XLM-RoBERTa may be beneficial for medical terminology and foreign entity forms, many of which cannot be directly factorized as native Vietnamese syllables.

\subsubsection{Social-media and feedback understanding}

\begin{table*}[htp]
    \centering
    \resizebox{\linewidth}{!}{
    \begin{tabular}{lcccccccc}
    \toprule
    \textbf{Model} & \textbf{VSMEC} & \textbf{ViHSD} & \textbf{ViSFD} & \textbf{ViOCD} & \multicolumn{2}{c}{\textbf{ViCTSD}} & \multicolumn{2}{c}{\textbf{UIT-VSFC}} \\
    \cmidrule(lr){6-7} \cmidrule(lr){8-9} & & & & & \textbf{Cons.} & \textbf{Toxic} & \textbf{Topic} & \textbf{Sent.} \\
    \midrule
    WikiBERT & 57.64 & -- & \underline{71.46} & -- & -- & -- & -- & -- \\
    PhoBERT & 59.91 & -- & -- & -- & \makecell{$\mathbf{82.11}$ \\ $\pm1.00$} & \makecell{$72.53$ \\ $\pm1.66$} & \makecell{$\underline{78.78}$ \\ $\pm0.71$} & \makecell{$\underline{80.95}$ \\ $\pm1.26$} \\
    mBERT & 54.59 & \underline{64.20} & 70.27 & \underline{91.61} & \makecell{$\underline{81.62}$ \\ $\pm0.72$} & \makecell{$68.53$ \\ $\pm3.67$} & \makecell{$75.80$ \\ $\pm0.71$} & \makecell{$78.30$ \\ $\pm1.46$} \\
    DistilBERT & 53.83 & 62.50 & 70.97 & 90.50 & -- & -- & -- & -- \\
    XLM-RoBERTa & \textbf{61.89} & 63.68 & -- & -- & \makecell{$\underline{81.62}$ \\ $\pm0.61$} & \makecell{$\underline{73.29}$ \\ $\pm2.09$} & \makecell{$77.66$ \\ $\pm1.79$} & \makecell{$80.57$ \\ $\pm1.72$} \\
    \midrule
    PhonemicBERT-Vi & \makecell{$\underline{60.79}$ \\ $\pm1.54$} & \makecell{$\mathbf{64.67}$ \\ $\pm0.31$} & \makecell{$\mathbf{75.36}$ \\ $\pm0.33$} & \makecell{$\mathbf{92.69}$ \\ $\pm1.11$} & \makecell{$81.59$ \\ $\pm1.04$} & \makecell{$\mathbf{79.88}$ \\ $\pm1.10$} & \makecell{$\mathbf{78.94}$ \\ $\pm0.25$} & \makecell{$\mathbf{82.21}$ \\ $\pm1.22$} \\
    \bottomrule
    \end{tabular}}
    \caption{Macro-F1 results on Vietnamese social-media and feedback benchmarks. ViCTSD includes constructiveness and toxicity detection, while UIT-VSFC includes topic and sentiment classification. Results on VSMEC, ViHSD, ViSFD, and ViOCD are reported from \cite{do-etal-2024-vlue}, while results on the ViCTSD and UIT-VSFC are from \cite{dinh-etal-2026-morphology}.}
    \label{tab:vi_social_results}
\end{table*}

Table~\ref{tab:vi_social_results} presents results on social-media classification, hate-speech detection, complaint detection, toxicity, topic, and sentiment analysis. PhonemicBERT-Vi achieves the highest reported mean on six of the eight prediction targets and ranks second on VSMEC.

PhonemicBERT-Vi obtains the highest reported results on ViHSD ($64.67\pm0.31$), UIT-ViSFD ($75.36\pm0.33$), and ViOCD ($92.69\pm1.11$). Its largest improvement among these datasets occurs on UIT-ViSFD, where it exceeds WikiBERT by 3.90 points. On ViOCD, it improves over mBERT by 1.08 points, while its 0.47-point advantage over mBERT on ViHSD is comparatively small.

The two multi-target datasets reveal different patterns. On ViCTSD, PhonemicBERT-Vi obtains $79.88\pm1.10$ on toxicity detection, 6.59 points above XLM-RoBERTa, the strongest reported baseline. On constructiveness, however, its score of $81.59\pm1.04$ is effectively comparable to PhoBERT, mBERT, and XLM-RoBERTa, whose reported means range from 81.62 to 82.11.

On UIT-VSFC, PhonemicBERT-Vi achieves the highest reported means for both topic classification ($78.94\pm0.25$) and sentiment classification ($82.21\pm1.22$). The differences from PhoBERT are only 0.16 and 1.26 points, respectively, so these results are best characterized as competitive rather than substantial improvements. On VSMEC, PhonemicBERT-Vi ranks second with $60.79\pm1.54$, behind XLM-RoBERTa at 61.89.

These results demonstrate that the complete phonemic pipeline remains effective on informal text despite the prevalence of teen code, emoticons, foreign expressions, and malformed forms. Such inputs invoke the character-level fallback mechanism rather than phonological factorization. The results therefore show operational robustness in the presence of fallback, but do not establish that fallback itself is responsible for the performance gains.

\subsubsection{Inference and question answering}

\begin{table}[th]
    \centering
    \small
    \begin{tabular}{lcc}
    \toprule
    \textbf{Model} & \textbf{ViNewsQA} & \textbf{ViNLI} \\
    \midrule
    WikiBERT & \underline{82.85} & -- \\
    PhoBERT & -- & \textbf{78.05} \\
    mBERT & \textbf{83.19} & 73.62 \\
    DistilBERT & -- & 66.77 \\
    XLM-RoBERTa & -- & \underline{77.01} \\
    \midrule
    PhonemicBERT-Vi & $82.16\pm1.35$ & $73.47\pm1.20$ \\
    \bottomrule
    \end{tabular}
    \caption{Results on Vietnamese inference and question-answering benchmarks. ViNewsQA is evaluated using extractive QA F1, while ViNLI uses macro-F1.}
    \label{tab:vi_reasoning_results}
\end{table}

Table~\ref{tab:vi_reasoning_results} reports results on ViNewsQA and ViNLI, which emphasize document-level question answering and sentence-pair semantic reasoning, respectively. On ViNewsQA, PhonemicBERT-Vi obtains $82.16\pm1.35$, compared with 83.19 for mBERT and 82.85 for WikiBERT. Its mean is 1.03 points below the strongest reported result, indicating broadly competitive question-answering performance. A larger deficit appears on ViNLI. PhonemicBERT-Vi obtains $73.47\pm1.20$, compared with 78.05 for PhoBERT and 77.01 for XLM-RoBERTa. Unlike POS tagging and NER, natural language inference depends less directly on local syllable alignment and more strongly on sentence-level lexical and semantic distinctions.

\subsubsection{Model and vocabulary scale}

\begin{table*}[thp]
    \centering
    \resizebox{\linewidth}{!}{
    \begin{tabular}{lrrrrrll}
    \toprule
    \textbf{Pretrained model} & \textbf{\#Params} & \textbf{\#Layers} & \textbf{\#Heads} & \textbf{Hidden} & \textbf{Vocab.} & \textbf{Language} & \textbf{Pretraining source} \\
    \midrule
    mBERT & 179M & 12 & 12 & 768 & 119,547 & Multilingual & Wikipedia \\
    Multilingual DistilBERT & 134M & 6 & 12 & 768 & 119,547 & Multilingual & Wikipedia \\
    XLM-RoBERTa-base & 270M & 12 & 12 & 768 & 250,002 & Multilingual & CommonCrawl \\
    WikiBERT & -- & 12 & 12 & 768 & 20,101 & Monolingual & Wikipedia \\
    PhoBERT-base & 135M & 12 & 12 & 768 & 64,001 & Monolingual & Wikipedia, news \\
    \midrule
    PhonemicBERT-Vi & \textbf{87.94M} & 12 & 12 & 768 & \textbf{256} & Monolingual & Wikipedia, news \\
    \bottomrule
    \end{tabular}}
    \caption{Architectural and pretraining characteristics of the Vietnamese and multilingual pretrained models. Parameter and vocabulary counts correspond to the evaluated checkpoints. A dash indicates that the parameter count is not reported.}
    \label{tab:vi_model_comparison}
\end{table*}

Table~\ref{tab:vi_model_comparison} contextualizes the downstream results with respect to model capacity, vocabulary size, and pretraining source. Phonemi  cBERT-Vi uses the same 12-layer depth, 768-dimensional hidden representations, and 12 attention heads as a standard base-sized encoder. However, its shared phonemic vocabulary contains only 256 entries, compared with 20,101--250,002 entries for the pretrained baselines.

PhonemicBERT-Vi consequently contains 87.94M parameters, approximately 34.9\% fewer than PhoBERT-base, 50.9\% fewer than mBERT, and 67.4\% fewer than XLM-RoBERTa-base. Although the factorized input projection and three H-MLM prediction heads introduce additional parameters, their cost remains substantially smaller than that of the large embedding and output matrices required by conventional token vocabularies.

\subsubsection{Overall findings} Across the three task groups, PhonemicBERT-Vi achieves the highest reported mean on nine of fourteen prediction targets. Its strongest results occur on token- and span-aligned tasks and on several social-media benchmarks, while its performance is less competitive on natural language inference. These results show that a model pretrained with the Phonemic Tokenizer can remain competitive with larger monolingual and multilingual PLMs, although its effectiveness varies across task types.

\subsection{Controlled Chinese Comparison} \label{sec:downstream_results_zh}

We compare PhonemicBERT-Zh with character, subword, and SubChar models under the Chinese experimental protocol of \citet{si2023subchar}. Results are organized into text classification, sentence-pair and reasoning, and reading comprehension and sequence-labeling tasks. We discuss the six- and twelve-layer configurations separately rather than relying on an aggregate score.

\subsubsection{Text classification}

\begin{table}[th]
    \centering
    \small
    \begin{tabular}{lccc}
    \toprule
    \textbf{Tokenizer} & \textbf{TNEWS} & \textbf{IFLYTEK} & \textbf{THUCNews} \\
    \midrule
    \multicolumn{4}{c}{\textit{6-layer encoder}} \\
    \midrule
    CharTokenizer & $\mathbf{64.19}\pm0.18$ & $55.83\pm0.50$ & $96.95\pm0.04$ \\
    Subword & $64.09\pm0.28$ & $54.88\pm0.39$ & $\mathbf{97.14}\pm0.03$ \\
    SubChar-Wubi & $63.89\pm0.25$ & $\underline{58.64}\pm0.27$ & $97.02\pm0.04$ \\
    SubChar-Pinyin & $63.68\pm0.25$ & $\mathbf{58.81}\pm0.28$ & $\underline{97.04}\pm0.04$ \\
    PhonemicBERT-Zh & $\underline{64.13}\pm0.21$ & $58.14\pm0.31$ & $97.02\pm0.04$ \\
    \midrule
    \multicolumn{4}{c}{\textit{12-layer encoder}} \\
    \midrule
    CharTokenizer & $\underline{64.39}\pm0.13$ & $58.52\pm0.46$ & $97.02\pm0.03$ \\
    SubChar-Pinyin & $64.19\pm0.14$ & $\underline{59.67}\pm0.23$ & $\underline{97.12}\pm0.03$ \\
    PhonemicBERT-Zh & $\mathbf{64.40}\pm0.12$ & $\mathbf{60.47}\pm0.25$ & $\mathbf{97.16}\pm0.04$ \\
    \bottomrule
    \end{tabular}
    \caption{Accuracy on Chinese text-classification benchmarks, reported as mean $\pm$ standard deviation. Bold and underlining indicate the highest and second-highest means within each encoder configuration.}
    \label{tab:zh_classification_results}
\end{table}

Table~\ref{tab:zh_classification_results} reports results on TNEWS, IFLYTEK, and THUCNews. With six layers, PhonemicBERT-Zh remains competitive on all three datasets. It obtains $64.13\pm0.21$ on TNEWS, only 0.06 points below CharTokenizer; $58.14\pm0.31$ on IFLYTEK, compared with $58.81\pm0.28$ for SubChar-Pinyin; and $97.02\pm0.04$ on THUCNews, compared with $97.14\pm0.03$ for Subword.

Scaling PhonemicBERT-Zh to twelve layers improves its mean on all three datasets. It obtains the highest reported means on TNEWS ($64.40\pm0.12$), IFLYTEK ($60.47\pm0.25$), and THUCNews ($97.16\pm0.04$). The clearest difference occurs on IFLYTEK, where it exceeds SubChar-Pinyin by 0.80 points. On TNEWS and THUCNews, the differences from the second-highest means are only 0.01 and 0.04 points, respectively, and the runs exhibit overlapping variation. These results should therefore be interpreted as competitive performance rather than substantial improvements.

Overall, phonemic factorization preserves text-classification performance despite reducing the vocabulary from 22,675 entries to 112. The larger gain on IFLYTEK may indicate that the deeper encoder is better able to contextualize factorized syllable representations over its comparatively long input descriptions, although the current experiments do not isolate this effect.

\subsubsection{Sentence-pair and reasoning tasks}

\begin{table*}[htp]
    \centering
    \resizebox{\linewidth}{!}{
    \setlength{\tabcolsep}{4.5pt}
    \begin{tabular}{lccccc}
    \toprule
    \textbf{Tokenizer} & \textbf{CLUEWSC} & \textbf{AFQMC} & \textbf{OCNLI} & \textbf{CSL} & \textbf{BQ} \\
    \midrule
    \multicolumn{6}{c}{\textit{6-layer encoder}} \\
    \midrule
    CharTokenizer & $63.39\pm1.95$ & $68.68\pm0.46$ & $68.19\pm0.39$ & $82.67\pm0.46$ & $\underline{81.99}\pm0.47$ \\
    Subword & $62.67\pm2.87$ & $\mathbf{69.25}\pm0.42$ & $\mathbf{69.03}\pm0.44$ & $\underline{83.20}\pm0.27$ & $81.94\pm0.28$ \\
    SubChar-Wubi & $64.61\pm2.09$ & $68.75\pm0.59$ & $\underline{68.93}\pm0.38$ & $82.81\pm0.46$ & $81.70\pm0.29$ \\
    SubChar-Pinyin & $\mathbf{65.90}\pm1.45$ & $68.89\pm0.42$ & $67.98\pm0.45$ & $82.87\pm0.40$ & $81.74\pm0.24$ \\
    PhonemicBERT-Zh & $\underline{65.17}\pm1.62$ & $\mathbf{69.25}\pm0.38$ & $67.73\pm0.40$ & $\mathbf{83.22}\pm0.23$ & $\mathbf{83.08}\pm0.27$ \\
    \midrule
    \multicolumn{6}{c}{\textit{12-layer encoder}} \\
    \midrule
    CharTokenizer & $68.09\pm1.59$ & $69.00\pm0.35$ & $\underline{70.40}\pm0.34$ & $\underline{82.77}\pm0.33$ & $\underline{83.49}\pm0.38$ \\
    SubChar-Pinyin & $\mathbf{71.71}\pm2.03$ & $\underline{69.30}\pm0.24$ & $\mathbf{70.43}\pm0.25$ & $82.23\pm0.27$ & $82.28\pm0.16$ \\
    PhonemicBERT-Zh & $\underline{68.47}\pm1.75$ & $\mathbf{69.81}\pm0.28$ & $69.15\pm0.27$ & $\mathbf{83.37}\pm0.35$ & $\mathbf{84.33}\pm0.24$ \\
    \bottomrule
    \end{tabular}}
    \caption{Accuracy on Chinese sentence-pair and reasoning benchmarks, reported as mean $\pm$ standard deviation.}
    \label{tab:zh_reasoning_results}
\end{table*}

Table~\ref{tab:zh_reasoning_results} presents results on CLUEWSC, AFQMC, OCNLI, CSL, and BQ. In the six-layer setting, PhonemicBERT-Zh ties Subword for the highest AFQMC mean at 69.25 and obtains the highest reported means on CSL ($83.22\pm0.23$) and BQ ($83.08\pm0.27$).

In the twelve-layer setting, PhonemicBERT-Zh achieves the highest reported means on AFQMC ($69.81\pm0.28$), CSL ($83.37\pm0.35$), and BQ ($84.33\pm0.24$). Its BQ mean exceeds CharTokenizer by 0.84 points, while its AFQMC and CSL means exceed the strongest baselines by 0.51 and 0.60 points, respectively. Together with the six-layer results, this pattern shows that the factorized representation remains effective for semantic equivalence and keyword--abstract consistency tasks.

The results are weaker on CLUEWSC and OCNLI. On twelve-layer CLUEWSC, PhonemicBERT-Zh obtains $68.47\pm1.75$, compared with $71.71\pm2.03$ for SubChar-Pinyin. On OCNLI, it reaches $69.15\pm0.27$, trailing SubChar-Pinyin by 1.28 points. The standard deviations indicate considerable run-to-run variation on CLUEWSC, whereas the OCNLI difference is more consistent.

Both tasks require fine-grained contextual distinctions, and phonemic factorization does not uniformly preserve every orthographic distinction available to character- or indexed Pinyin-based representations. In particular, different Chinese characters can map to the same phonemic tuple. We investigate the relationship between such homophonic collisions and downstream behavior in Section~\ref{sec:homophone_analysis}.

\subsubsection{Reading comprehension and sequence labeling}

\begin{table}[th]
    \centering
    \small
    \begin{tabular}{lccc}
    \toprule
    \textbf{Tokenizer} & \textbf{C3} & \textbf{CMRC}$^{\dagger}$ & \textbf{CLUENER}$^{\dagger}$ \\
    \midrule
    \multicolumn{4}{c}{\textit{6-layer encoder}} \\
    \midrule
    CharTokenizer & $53.17\pm0.56$ & \textbf{56.58} & 69.61 \\
    Subword & $53.32\pm0.44$ & \underline{55.85} & 67.94 \\
    SubChar-Wubi & $\mathbf{54.68}\pm0.77$ & 54.45 & 70.63 \\
    SubChar-Pinyin & $53.03\pm0.47$ & 55.18 & \textbf{70.77} \\
    PhonemicBERT-Zh & $\underline{53.69}\pm0.56$ & $55.49\pm0.11$ & $\underline{70.71}\pm0.03$ \\
    \midrule
    \multicolumn{4}{c}{\textit{12-layer encoder}} \\
    \midrule
    CharTokenizer & $54.22\pm0.40$ & -- & -- \\
    SubChar-Pinyin & $\mathbf{55.92}\pm0.26$ & -- & -- \\
    PhonemicBERT-Zh & $\underline{55.33}\pm0.21$ & -- & -- \\
    \bottomrule
    \end{tabular}
    \caption{Results on Chinese reading-comprehension and sequence-labeling benchmarks. C3 is evaluated using accuracy, CMRC using exact match (EM), and CLUENER using entity-level F1. All results are reported as mean $\pm$ standard deviation except the published CMRC and CLUENER baselines marked with $\dagger$, for which \citet{si2023subchar} report single-run results. Twelve-layer CMRC and CLUENER baseline results are unavailable.}
    \label{tab:zh_reading_results}
\end{table}

Table~\ref{tab:zh_reading_results} presents task-standard scores on C3, CMRC, and CLUENER. C3 is a multiple-choice reading-comprehension benchmark evaluated using accuracy, whereas CMRC is a span-extraction benchmark evaluated using exact match. CLUENER is evaluated using entity-level F1.

In the six-layer setting, PhonemicBERT-Zh obtains $53.69\pm0.56$ accuracy on C3, ranking second behind SubChar-Wubi at $54.68\pm0.77$. Scaling to twelve layers improves its C3 accuracy to $55.33\pm0.21$, leaving it 0.59 points below SubChar-Pinyin.

On CMRC, PhonemicBERT-Zh obtains $55.49\pm0.11$ EM, compared with the single-run CharTokenizer result of 56.58 and Subword result of 55.85. On CLUENER, it reaches $70.71\pm0.03$ entity-level F1, closely matching the single-run SubChar-Pinyin score of 70.77.

\subsubsection{Analysis on Homophone in Chinese} \label{sec:homophone_analysis}

\begin{figure}[htp]
    \centering
    \includegraphics[width=0.75\linewidth]{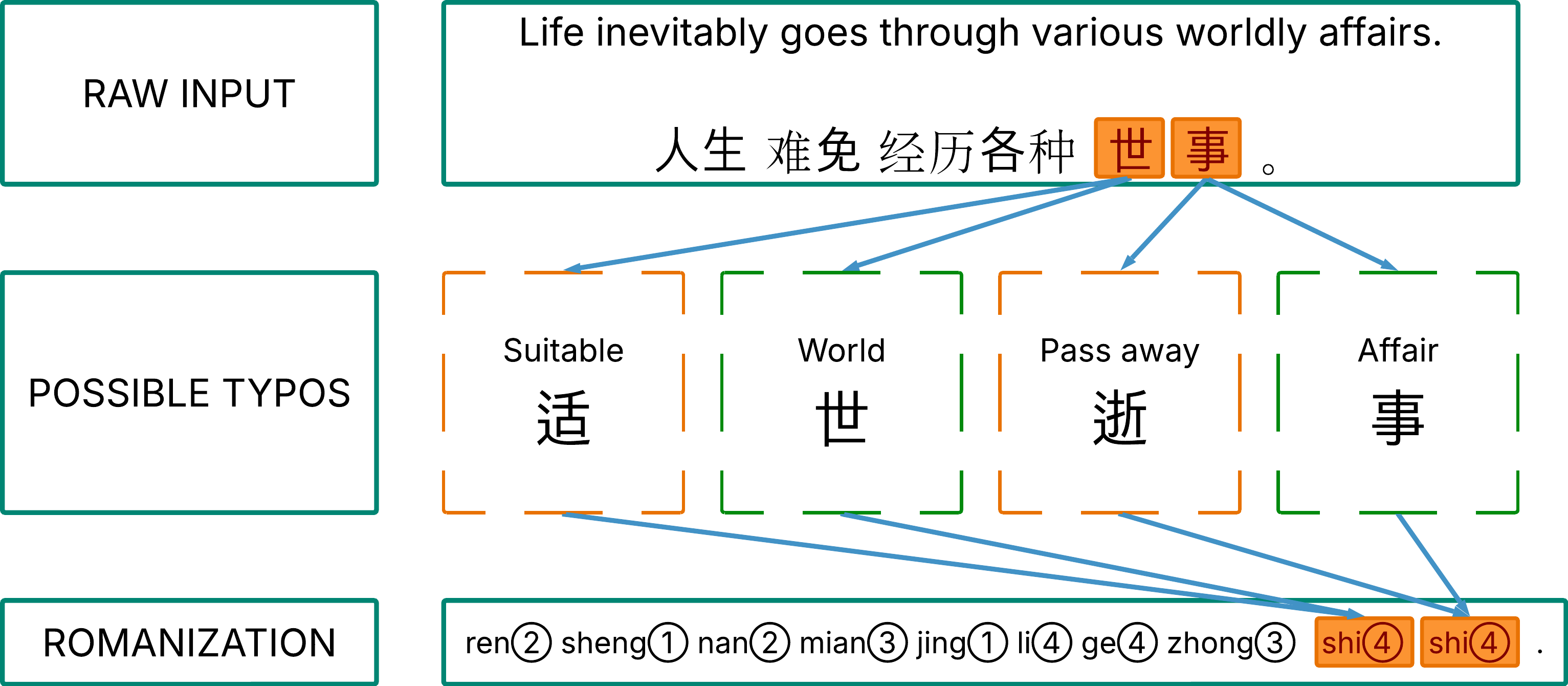}
    \caption{Illustration of synthetic homophone generation \cite{si2023subchar}.}
    \label{fig:homophone}
\end{figure}

A phonemic representation maps homophonous characters to the same pronunciation tuple. This property may remove orthographic distinctions between characters with different meanings, but it should also make the tokenizer robust to homophone substitutions, a common category of Chinese input errors. We therefore investigate how synthetic homophones affect PhonemicBERT-Zh.

We follow the evaluation procedure of SubChar \cite{si2023subchar}. Given a clean input sequence, a specified proportion of Chinese characters is randomly selected and replaced with different characters having the same Pinyin pronunciation, including lexical tone. The replacement character is therefore orthographically and potentially semantically different from the original character, while preserving its pronunciation. For example, the characters "世" (world) and "事" (affair) in "世事" (\textit{shìshì}) can be replaced by the homophones "适" (suitable) and "逝" (pass away), respectively, because all four characters are pronounced as \textit{shi\textcircled{4}} (Figure \ref{fig:homophone}). We evaluate corruption ratios of $7.5\%$, $15\%$, $22.5\%$, $30\%$, and $37.5\%$.

This procedure directly tests whether the model's input representation is stable under pronunciation-preserving orthographic perturbations. Character and conventional subword tokenizers assign different token identities to the original and substituted characters. By contrast, Phonemic Tokenizer factorizes homophones into the same onset, rime, and tone components when their assigned pronunciations are identical. We conduct the analysis on TNEWS, OCNLI, and C3 and report accuracy. Results for the Character, Subword, SubChar-Pinyin, and SubChar-Pinyin-NoIndex baselines are taken from \citet{si2023subchar}.

\begin{table*}[ht]
    \centering
    \resizebox{\linewidth}{!}{
    \begin{tabular}{llrrrrrr}
    \toprule
    \textbf{Dataset} & \textbf{Tokenizer} & \textbf{Clean} & \textbf{7.5\%} & \textbf{15.0\%} & \textbf{22.5\%} & \textbf{30.0\%} & \textbf{37.5\%} \\
    \midrule
    
    \multirow{5}{*}{TNEWS}
    & Character              & 64.10 & 63.09 & 58.96 & 50.91 & 38.33 & 25.20 \\
    & Subword                & 64.09 & 62.82 & 57.75 & 48.67 & 36.37 & 25.27 \\
    & SubChar-Pinyin         & 63.68 & 61.95 & 56.67 & 45.22 & 30.71 & 27.53 \\
    & SubChar-Pinyin-NoIndex & 63.28 & 63.28 & \textbf{63.28} & \textbf{63.28} & \textbf{63.28} & 63.28 \\
    & PhonemicBERT-Zh        & \textbf{64.07} & \textbf{63.71} & 63.17 & 63.10
                                          & 63.09 & \textbf{63.39} \\
    \midrule
    
    \multirow{5}{*}{OCNLI}
    & Character              & 68.19 & 64.89 & 56.58 & 47.65 & 40.48 & 36.36 \\
    & Subword                & \textbf{69.03} & 64.33 & 56.49 & 48.07 & 42.68 & 38.28 \\
    & SubChar-Pinyin         & 67.70 & 61.93 & 54.39 & 46.01 & 40.24 & 37.33 \\
    & SubChar-Pinyin-NoIndex & 67.91 & \textbf{67.91} & \textbf{67.91} & \textbf{67.91} & \textbf{67.91} & \textbf{67.91} \\
    & PhonemicBERT-Zh        & 67.32 & 66.88 & 66.61 & 67.22 & 66.92 & 67.12 \\
    \midrule
    
    \multirow{5}{*}{C3}
    & Character              & 53.13 & 51.46 & 49.22 & 47.71 & 46.78 & 43.95 \\
    & Subword                & 53.55 & 51.66 & 49.49 & 47.81 & 46.24 & 43.58 \\
    & SubChar-Pinyin         & 52.87 & 50.45 & 47.46 & 44.50 & 42.42 & 40.07 \\
    & SubChar-Pinyin-NoIndex & 53.65 & \textbf{53.65} & \textbf{53.65} & \textbf{53.65} & \textbf{53.65} & \textbf{53.65} \\
    & PhonemicBERT-Zh        & \textbf{53.70} & 52.15 & 52.04 & 52.18 & 51.81 & 51.81 \\
    \bottomrule
    \end{tabular}}
    \caption{Accuracy under synthetic homophone substitutions. Corruption percentages indicate the proportion of Chinese characters replaced with different characters having the same Pinyin pronunciation and tone. Baseline results are taken from \citet{si2023subchar}. Bold value indicate the best result within the pronunciation-based representations for each setting.}
    \label{tab:zh_homophone_robustness}
\end{table*}

Table~\ref{tab:zh_homophone_robustness} shows a sharp contrast between orthography-dependent and pronunciation-based representations. Character, Subword, and SubChar-Pinyin models deteriorate progressively as the corruption ratio increases. At $37.5\%$ corruption, the Character model drops by 38.90 points on TNEWS, 31.83 points on OCNLI, and 9.18 points on C3. The corresponding decreases for the Subword model are 38.82, 30.75, and 9.97 points. Applying BPE after Pinyin conversion does not eliminate this sensitivity: SubChar-Pinyin loses 36.15, 30.37, and 12.80 points on the three datasets, respectively. Statistical segmentation can change when characters are substituted even if their pronunciations are preserved.

PhonemicBERT-Zh is substantially more stable. From the clean condition to $37.5\%$ corruption, its accuracy changes from 64.07 to 63.39 on TNEWS, from 67.32 to 67.12 on OCNLI, and from 53.70 to 51.81 on C3, corresponding to decreases of only 0.68, 0.20, and 1.89 points. Thus, even at the highest corruption level, the model retains nearly all of its clean-input performance on TNEWS and OCNLI and most of its performance on C3. These results confirm that phonological factorization provides robustness to orthographic substitutions that preserve pronunciation.

SubChar-Pinyin-NoIndex is exactly invariant across corruption levels because homophonous characters are deliberately mapped to the same unindexed romanized representation. PhonemicBERT-Zh exhibits a similar, although not perfect, invariance. This difference is attributable to the phrase-aware Chinese pronunciation pipeline. Phonemic Tokenizer applies longest-match-first lookup to assign contextual pronunciations; replacing a character can break or create a phrase match, thereby changing the pronunciation assigned to that position or a neighboring position even when the substituted character is a homophone in isolation. This effect explains the small fluctuations observed across corruption ratios. It is most visible on C3, where longer and more context-dependent inputs provide more opportunities for phrase matches to change.

Overall, these results reveal both the strength and the limitation of the Phonemic Tokenizer. By mapping pronunciation-preserving substitutions to identical or highly similar phonemic tuples, it substantially improves robustness to homophone-based noise. However, the same factorization removes the orthographic identity that distinguishes homophonous characters with different meanings. Such characters become indistinguishable at the input level, requiring PhonemicBERT-Zh to recover their semantic distinctions entirely from the surrounding context. Phonemic Tokenizer therefore trades character-level lexical specificity for greater representational sharing and robustness to orthographic variation.

\section{Conclusion} \label{sec:conclusion}

We introduced the \textbf{Phonemic Tokenizer}, a linguistically motivated alternative to character and statistical subword tokenization for Vietnamese and Chinese. Instead of treating a syllable as an atomic symbol or fragmenting it into statistically derived pieces, the proposed tokenizer factorizes each syllable into three phonological components---onset, rime, and tone---while preserving one contextual position per syllable. Its deterministic IPA-based formulation requires no corpus-dependent vocabulary learning and represents the phonological systems of Chinese and Vietnamese with compact vocabularies of 112 and 256 entries, respectively.

Our intrinsic evaluation shows that this factorization achieves consistently high Rényi efficiency in both languages. In Vietnamese, it represents every entry in the standard syllable dictionary with a Fertility of exactly one and generally produces shorter sequences than the evaluated pretrained tokenizers. In Chinese, it preserves character-level sequence length while using a substantially smaller vocabulary than character, subword, and SubChar alternatives. These findings demonstrate that compactness need not be obtained by increasing sequence length or fragmenting syllables into multiple contextual positions.

To evaluate whether the proposed representation remains effective for language modeling, we instantiated it in \textbf{PhonemicBERT}, which combines the three component embeddings through a learned projection and reconstructs complete masked syllables using three prediction heads. PhonemicBERT-Zh remains competitive with character, subword, and SubChar models under a controlled Chinese pretraining setup, while PhonemicBERT-Vi outperforms established monolingual and multilingual pretrained models on several Vietnamese downstream tasks despite its much smaller vocabulary and parameter count. The synthetic homophone experiments further demonstrate that collapsing pronunciation-equivalent characters into shared phonemic representations yields stable downstream performance under homophone substitutions, although it also removes orthographic distinctions among semantically different homophones.

\section*{Acknowledgement}
This research is funded by Vietnam National University Ho Chi Minh City (VNU-HCM) under grant number NCM2025-26-02. This work also forms part of the doctoral dissertation of the PhD candidate identified by student ID NCS26006.


\clearpage

\bibliographystyle{elsarticle-num-names} 
\bibliography{main}

\clearpage

\appendix

\section{Vietnamese Phonology and Orthography} \label{sec:app_vi}

This appendix summarizes the syllable structure of Vietnamese and the systematic relationship between its modern orthography and phonological representation. These linguistic properties provide the basis for the deterministic factorization performed by Phonemic Tokenizer.

\subsection{Syllable Structure}

Vietnamese belongs to the Vietic branch of the Austroasiatic language family. Modern Vietnamese is written in \textit{Chữ Quốc Ngữ}, a Latin based orthography developed to represent Vietnamese pronunciation \cite{haophonetic,caophoneme,hao1998,thuat2016}. In this writing system, whitespace generally separates orthographic syllables rather than necessarily delimiting complete lexical words. A multisyllabic Vietnamese word may therefore consist of multiple whitespace-separated orthographic units.

Phonologically, a Vietnamese syllable follows the canonical structure
\begin{equation}
    S = (O,R,T),
    \label{eq:vi_syllable_structure}
\end{equation}
where $O$, $R$, and $T$ denote the onset, rime, and lexical tone, respectively. The onset is optional, whereas the rime constitutes the obligatory core of the syllable. The rime can be further analyzed as
\begin{equation}
    R = (G,V,F),
    \label{eq:vi_rime_structure}
\end{equation}
where $G$ is an optional medial glide, $V$ is the vocalic nucleus, and $F$ is an optional coda.

For example, under our composite pronunciation standard, the orthographic syllable \textit{hoàng} is transcribed as

\begin{center}
    \textit{hoàng} $\longmapsto$/\textipa{h\textsubarch{u}a\ng\tone{21}}/.
\end{center}
Its IPA representation can be analyzed as 
\begin{center}
    O = /\textipa{h}/, \\
    G = /\textipa{\textsubarch{u}}/, \\
    V = /\textipa{a}/, \\
    F = /\textipa{\ng}/, \\
    T = /\textipa{\tone{21}}/
\end{center}
The medial, nucleus, and coda are then combined into the atomic rime
\begin{center}
    R = /\textipa{wa\ng}/
\end{center}
giving the final phonological factorization
\begin{center}
    \textit{hoàng} $\longmapsto$ (/\textipa{h}/, /\textipa{wa\ng}/, /\textipa{\tone{21}}/).
\end{center}


Vietnamese distinguishes six lexical tones: \textit{ngang} /\textipa{\tone{33}}/, \textit{huyền} /\textipa{\tone{22}\tone{11}}/, \textit{sắc} /\textipa{\tone{33}\tone{55}}/, \textit{hỏi} /\textipa{\tone{33}\tone{11}}/, \textit{ngã} /\textipa{\tone{33}P\tone{55}}/, and \textit{nặng} /\textipa{\tone{33}P\tone{11}}/. They form an integral part of the syllable and can distinguish lexical meaning.

\subsection{Correspondence between Orthography and Phonology}

\textit{Chữ Quốc Ngữ} exhibits a high degree of correspondence between graphemes and phonological categories. Given a well-formed Vietnamese syllable, its pronunciation can generally be determined from its spelling using a finite set of context-sensitive rules. This correspondence remains stable across grammatical contexts because Vietnamese words do not undergo the extensive inflectional changes found in many morphologically rich languages.

\begin{table*}[ht]
    \centering
    \resizebox{0.85\linewidth}{!}{
    \begin{tabular}{llll}
    \toprule
    \textbf{Grapheme(s)} & \textbf{Phoneme} & \textbf{Example} & \textbf{Example IPA} \\
    \midrule
    \textit{b}  & /\textipa{b}/ & \textit{ba}   & /\textipa{ba\tone{33}}/ \\
    \textit{t}  & /\textipa{t}/ & \textit{ta}   & /\textipa{ta\tone{33}}/ \\
    \textit{th} & /\textipa{t\super h}/ & \textit{tha}  & /\textipa{t\super ha\tone{33}}/ \\
    \textit{c}, \textit{k}, \textit{q} & /\textipa{k}/ & \textit{ca}, \textit{kim}, \textit{qua} & /\textipa{ka\tone{33}}/, /\textipa{kim\tone{33}}/, /\textipa{kwa\tone{33}}/ \\
    \textit{ph} & /\textipa{f}/ & \textit{pha}  & /\textipa{fa\tone{33}}/ \\
    \textit{đ}  & /\textipa{d}/ & \textit{đa}   & /\textipa{da\tone{33}}/ \\
    \textit{g}, \textit{gh} & /\textipa{7}/ & \textit{ga}, \textit{ghe} & /\textipa{7a\tone{33}}/, /\textipa{7E\tone{33}}/ \\
    \textit{gi} & /\textipa{z}/ & \textit{gia}  & /\textipa{za\tone{33}}/ \\
    \textit{d}  & /\textipa{j}/ & \textit{da}   & /\textipa{ja\tone{33}}/ \\
    \textit{x}  & /\textipa{s}/ & \textit{xa}   & /\textipa{sa\tone{33}}/ \\
    \textit{s}  & /\textipa{\:s}/ & \textit{sa}   & /\textipa{\:sa\tone{33}}/ \\
    \textit{ch} & /\textipa{\t{cC}}/ & \textit{cha}  & /\textipa{\t{cC}a\tone{33}}/ \\
    \textit{tr} & /\textipa{\t{t\:s}}/ & \textit{tra}  & /\textipa{\t{t\:s}a\tone{33}}/ \\
    \textit{ng}, \textit{ngh} & /\textipa{\ng}/ & \textit{nga}, \textit{nghe} & /\textipa{\ng a\tone{33}}/, /\textipa{\ng E\tone{33}}/ \\
    \textit{nh} & /\textipa{\textltailn}/ & \textit{nha}  & /\textipa{\textltailn a\tone{33}}/ \\
    \textit{l}  & /\textipa{l}/ & \textit{la}   & /\textipa{la\tone{33}}/ \\
    \textit{r}  & /\textipa{r}/ & \textit{ra}   & /\textipa{ra\tone{33}}/ \\
    \textit{kh} & /\textipa{x}/ & \textit{kha}  & /\textipa{xa\tone{33}}/ \\
    \textit{v}  & /\textipa{v}/ & \textit{va}   & /\textipa{va\tone{33}}/ \\
    \textit{m}  & /\textipa{m}/ & \textit{ma}   & /\textipa{ma\tone{33}}/ \\
    \textit{n}  & /\textipa{n}/ & \textit{na}   & /\textipa{na\tone{33}}/ \\
    \textit{h}  & /\textipa{h}/ & \textit{ha}   & /\textipa{ha\tone{33}}/ \\
    \bottomrule
    \end{tabular}}
    \caption{Orthographic onset-to-phoneme correspondences under the composite Vietnamese pronunciation standard. Tone contours in the examples describe phonetic realizations}
    \label{tab:vi_onset_mapping}
\end{table*}

The correspondence is not strictly one-to-one at the level of individual letters (Table \ref{tab:vi_onset_mapping}, \ref{tab:vi_vowel_mapping}, \ref{tab:vi_coda_mapping}, and \ref{tab:vi_tone_mapping}). A phoneme may be represented by different graphemes depending on the neighboring vowel or its position within the syllable. Nevertheless, these alternations are systematic rather than arbitrary. Representative cases include:

\begin{itemize}
    \item The same velar onset /\textipa{k}/ is written as \textit{c}, \textit{k}, or as part of \textit{qu}. The spelling \textit{k} is typically used before front vowels, \textit{qu} introduces a following medial glide, and \textit{c} occurs in the remaining environments.

    \item The spelling pairs \textit{g}/\textit{gh} and \textit{ng}/\textit{ngh} represent the same respective onset phonemes /\textipa{7}/ and /\textipa{\ng}/. The forms \textit{gh} and \textit{ngh} occur before particular front-vowel spellings, whereas \textit{g} and \textit{ng} are used elsewhere.

    \item A medial glide /\textipa{\textsubarch{u}}/ may be written as either \textit{o} or \textit{u}, with its orthographic realization determined by the surrounding onset and nucleus.

    \item The diphthong /\textipa{ie}/ represented phonologically as a single nucleus may appear as \textit{iê}, \textit{yê}, \textit{ia}, or \textit{ya}. The selected spelling depends on whether the syllable contains a medial or a following coda.

    \item The diphthong families /\textipa{uo}/ (\textit{uô}-\textit{ua}) and /\textipa{W9}/ (\textit{ươ}-\textit{ưa}) exhibit similar alternations. Forms containing \textit{ô} or \textit{ơ} occur before a coda, whereas \textit{ua} or \textit{ưa} occurs in open syllables.

    \item The vowel and semivowel spellings \textit{i}-\textit{y} /\textipa{\textsubarch{i}}/ and \textit{o}-\textit{u} /\textipa{\textsubarch{u}}/ are conditioned by their syllabic positions and neighboring phonological components.
\end{itemize}

For example, the initial velar stop /\textipa{k}/ has different surface spellings in \textit{ca} /\textipa{ka}/, \textit{kim} /\textipa{kim}/, and \textit{qua} /\textipa{k\textsubarch{u}a}/, but these forms can be normalized to the same onset category. Similarly, the nuclei in \textit{chia} /\textipa{\t{cC}ie}/, \textit{kiên} /\textipa{kien}/, \textit{yên} /\textipa{ien}/, and \textit{khuya} /\textipa{x\textsubarch{u}ie}/ exhibit different spellings whose selection is predictable from the surrounding syllable structure.

\begin{table*}[ht]
    \centering
    \resizebox{0.75\linewidth}{!}{
    \begin{tabular}{llll}
    \toprule
    \textbf{Component} & \textbf{Grapheme(s)} & \textbf{Phoneme} & \textbf{Examples} \\
    \midrule
    Medial & \textit{o}, \textit{u} & /\textipa{\textsubarch{u}}/ & \textit{hoa}, \textit{thuê}, \textit{khuyên} \\
    \midrule
    Monophthong & \textit{a} & /\textipa{a}/ & \textit{ba}, \textit{an} \\
    Monophthong & \textit{ă}; \textit{a} before \textit{y, u} & /\textipa{\u{a}}/ & \textit{ăn}, \textit{tay}, \textit{cau} \\
    Monophthong & \textit{â} & /\textipa{\u{9}}/ & \textit{ân}, \textit{cây} \\
    Monophthong & \textit{i}, \textit{y} & /\textipa{i}/ & \textit{tin}, \textit{y} \\
    Monophthong & \textit{ê} & /\textipa{e}/ & \textit{êm}, \textit{kết} \\
    Monophthong & \textit{e} & /\textipa{E}/ & \textit{em}, \textit{sen} \\
    Monophthong & \textit{u} & /\textipa{u}/ & \textit{thu}, \textit{ung} \\
    Monophthong & \textit{ư} & /\textipa{W}/ & \textit{tư}, \textit{ưng} \\
    Monophthong & \textit{o} & /\textipa{O}/ & \textit{ong}, \textit{con} \\
    Monophthong & \textit{oo} & /\textipa{O:}/ & \textit{xoong} \\
    Monophthong & \textit{ô} & /\textipa{o}/ & \textit{ông}, \textit{tốt} \\
    Monophthong & \textit{ơ} & /\textipa{9}/ & \textit{ơ}, \textit{sơn} \\
    \midrule
    Diphthong & \textit{iê}, \textit{yê}, \textit{ia}, \textit{ya} & /\textipa{ie}/ & \textit{biên}, \textit{yên}, \textit{bia}, \textit{khuya} \\
    Diphthong & \textit{uô}, \textit{ua} & /\textipa{uo}/ & \textit{buôn}, \textit{mua} \\
    Diphthong & \textit{ươ}, \textit{ưa} & /\textipa{W9}/ & \textit{hương}, \textit{mưa} \\
    \bottomrule
    \end{tabular}}
    \caption{Orthographic correspondences for the Vietnamese medial glide and vocalic nuclei. Multiple spellings in the same row are normalized to the same phonemic category.}
    \label{tab:vi_vowel_mapping}
\end{table*}

The relationship between orthography and pronunciation is therefore better characterized as \textbf{highly regular and context-conditioned} than as perfectly one-to-one. Several orthographic forms may map to the same phonological component, but their mappings can be resolved using the internal structure of the syllable. Consequently, Phonemic Tokenizer can
normalize surface spelling variants into canonical onset, rime, and tone components without learning a statistical segmentation model from a pretraining corpus.

This regularity also explains why phonological factorization does not require Vietnamese graphemes to be emitted as separate sequential tokens. Once an orthographic syllable has been analyzed, its internal phonological structure can be represented as a single aligned tuple:
\begin{equation}
    c_i \longmapsto (O_i,R_i,T_i),
\end{equation}
thereby preserving one contextual position per orthographic syllable while making its internal structure explicit.

\begin{align}
    \textit{bia} &\longmapsto/\textipa{bie\tone{33}}/, & \textit{biên} &\longmapsto /\textipa{bien\tone{33}}/, \\
    \textit{khuya} &\longmapsto/\textipa{x\textsubarch{u}ie\tone{33}}/, & \textit{khuyên} &\longmapsto /\textipa{x\textsubarch{u}ien\tone{33}}/.
\end{align}

\begin{table}[t]
    \centering
    \resizebox{0.55\linewidth}{!}{
    \begin{tabular}{lll}
    \toprule
    \textbf{Grapheme(s)} & \textbf{Phoneme} & \textbf{Examples} \\
    \midrule
    \textit{i}, \textit{y} & /\textipa{\textsubarch{i}}/ & \textit{mai}, \textit{tay} \\
    \textit{o}, \textit{u} & /\textipa{\textsubarch{u}}/ & \textit{sao}, \textit{đau} \\
    \textit{m} & /\textipa{m}/ & \textit{nam}, \textit{êm} \\
    \textit{n} & /\textipa{n}/ & \textit{an}, \textit{ên} \\
    \textit{ng} & /\textipa{\ng}/ & \textit{ang}, \textit{ông} \\
    \textit{nh} & /\textipa{\textltailn}/ & \textit{anh}, \textit{ênh} \\
    \textit{p} & /\textipa{p}/ & \textit{áp}, \textit{hợp} \\
    \textit{t} & /\textipa{t}/ & \textit{át}, \textit{kết} \\
    \textit{c} & /\textipa{k}/ & \textit{ác}, \textit{ức} \\
    \textit{ch} & /\textipa{c}/ & \textit{ách}, \textit{ích} \\
    \bottomrule
    \end{tabular}}
    \caption{Orthographic coda-to-phoneme correspondences used in the Vietnamese pronunciation normalization.}
    \label{tab:vi_coda_mapping}
\end{table}

\begin{table}[tp]
    \centering
    \small
    \begin{tabular}{lll}
    \toprule
    \textbf{Tone} & \textbf{Mark} & \textbf{Example} \\
    \midrule
    Ngang & Unmarked & \textit{ma} \\
    Huyền & Grave & \textit{mà} \\
    Sắc & Acute & \textit{má} \\
    Hỏi & Hook above & \textit{mả} \\
    Ngã & Tilde & \textit{mã} \\
    Nặng & Dot below & \textit{mạ} \\
    \bottomrule
    \end{tabular}
    \caption{Mapping from Vietnamese orthographic tone marks to the tone representations used by Phonemic Tokenizer.}
    \label{tab:vi_tone_mapping}
\end{table}

\section{Mandarin Chinese Phonology and Pronunciation Representation} \label{sec:app_zh}

This appendix describes the syllable structure of Mandarin Chinese and the relationship among Chinese characters, Pinyin, and IPA. Unlike Vietnamese \textit{Chữ Quốc Ngữ}, Chinese orthography does not transparently encode pronunciation. Phonemic Tokenizer therefore uses Pinyin as an intermediate pronunciation-bearing representation before converting each syllable into its IPA-based onset--rime--tone factorization.

\subsection{Mandarin Syllable Structure}

Modern Standard Mandarin is a tonal and predominantly syllable-based language. In most cases, one Chinese character corresponds to one spoken syllable, although the same character may have different pronunciations in different lexical contexts. Each Mandarin syllable is analyzed using the canonical structure
\begin{equation}
    S = (O,R,T),
    \label{eq:zh_syllable_structure}
\end{equation}
where $O$, $R$, and $T$ denote the onset, rime, and lexical tone, respectively. The onset is optional, whereas the rime forms the obligatory core of the syllable. Internally, a Mandarin rime may be analyzed as
\begin{equation}
    R = (G,V,F),
    \label{eq:zh_rime_structure}
\end{equation}
where $G$ denotes an optional medial glide, $V$ denotes the vocalic nucleus, and $F$ denotes an optional coda. As in the Vietnamese instantiation, these internal elements are used only to determine the rime. Phonemic Tokenizer stores their combination as one atomic rime component.

For example, the character \textit{光} (light) is represented in Pinyin as
\textit{guāng} and transcribed as
\begin{center}
    \textit{光} $\longmapsto$ \textit{guāng} $\longmapsto$ /\textipa{k\textsubarch{u}a\ng\tone{55}}/.
\end{center}
Its pronunciation can be analyzed as
\begin{center}
    O  = /\textipa{k}/, \\
    G = /\textipa{\textsubarch{u}}/, \\
    V = /\textipa{a}/, \\
    F = /\textipa{\ng}/, \\
    T = /\textipa{\tone{55}}/.
\end{center}
The glide, nucleus, and coda are combined into the atomic rime
\begin{center}
    R = /\textipa{\textsubarch{u}a\ng}/,
\end{center}
giving the final factorization
\begin{center}
    \textit{光} $\longmapsto$ (/\textipa{k}/, /\textipa{\textsubarch{u}a\ng}/, /\textipa{\tone{55}}/).
\end{center}

The same three-slot structure also represents syllables without an overt onset. For example, the character \textit{安} (peaceful), written as \textit{ān} in Pinyin and transcribed as /\textipa{an\tone{55}}/, is factorized as
\begin{center}
    \textit{安} $\longmapsto$ (\texttt{[EMPTY]}, /\textipa{an}/, /\textipa{\tone{55}}/).
\end{center}
Here, \texttt{[EMPTY]} represents the absence of an overt consonantal onset and is distinct from \texttt{[UNK]} and \texttt{[PAD]}.

Mandarin distinguishes four principal lexical tones, conventionally represented in Pinyin using tone marks or the numerical labels 1--4. The corresponding normalized pitch contours are high-level, rising, dipping, and falling. A neutral-tone category additionally occurs in unstressed syllables. The tone representation used by the tokenizer is assigned after phrase-level pronunciation retrieval, so the same character may receive different tone components in different lexical expressions.

\subsection{Relationship among Chinese Characters, Pinyin, and IPA}

Chinese characters primarily encode lexical and morphemic identity rather than directly specifying pronunciation. Consequently, the mapping between characters and spoken syllables is neither one-to-one nor recoverable from character shape alone. Two forms of ambiguity are particularly relevant.

First, many orthographically distinct characters are homophones. For example, \textit{意} (intention), \textit{义} (meaning), \textit{异} (strange), and \textit{议} (debate) can all be pronounced as \textit{yì}. Once converted into IPA, they share the same phonemic factorization:
\begin{center}
    \{\textit{意},\textit{义},\textit{异},\textit{议}\} $\longmapsto$ \textit{yì} $\longmapsto$ /\textipa{i\tone{51}}/.
\end{center}
This many-to-one mapping enables representation sharing and robustness to homophone substitutions, but it also removes the orthographic distinction among characters with different meanings.

Second, a single character may be polyphonic. For example, \textit{行} is pronounced as \textit{xíng} in \textit{行为} (behavior) but as \textit{háng} in \textit{银行} (bank):
\begin{align}
    \textit{行为} &\longmapsto \textit{xíngwéi}, \\
    \textit{银行} &\longmapsto \textit{yínháng}.
\end{align}
Pronunciation assignment must therefore precede IPA conversion and cannot always be performed independently for each character.

Pinyin bridges this gap by explicitly encoding the pronunciation of a Chinese syllable through an initial, a final, and a tone mark. Once the appropriate Pinyin reading has been selected, its conversion into IPA is highly systematic. We consequently decompose Chinese conversion into two functions:
\begin{equation}
    \psi_{\mathrm{Zh}}: C \longmapsto Q,
\end{equation}
where $\psi_{\mathrm{Zh}}$ assigns a context-sensitive Pinyin sequence $Q=[q_1,\ldots,q_n]$ to a character sequence, and
\begin{equation}
    \rho_{\mathrm{Zh}}: q_i \longmapsto s_i,
\end{equation}
where $\rho_{\mathrm{Zh}}$ deterministically converts a selected Pinyin syllable into its normalized IPA representation. The complete conversion is therefore
\begin{equation}
    \phi_{\mathrm{Zh}} = \rho_{\mathrm{Zh}} \circ \psi_{\mathrm{Zh}}.
    \label{eq:zh_complete_conversion}
\end{equation}

\subsection{Systematic Pinyin-to-IPA Conversion}

Although the relationship between Chinese characters and pronunciation is opaque, the relationship between Pinyin and IPA is highly regular. Pinyin letters should not, however, always be interpreted as independent Latin phonemes. Their pronunciation can depend on their position and neighboring letters within the syllable.

For example, the Pinyin forms \textit{gā} and \textit{gē} are transcribed under our normalization as
\begin{center}
    \textit{gā} $\longmapsto$ /\textipa{ka\tone{55}}/, \\
    \textit{gē} $\longmapsto$ /\textipa{k@\tone{55}}/.
\end{center}
Although their rimes differ, the Pinyin initial \textit{g} corresponds to the same normalized onset $/\textipa{k}/$ in both syllables.

Pinyin also employs the letters \textit{y} and \textit{w} as orthographic devices at the beginning of syllables that lack a consonantal initial. For example,
\begin{center}
    \textit{衣}~(\textit{yī}) $\longmapsto$ /\textipa{i\tone{55}}/, \\
    \textit{乌}~(\textit{wū}) $\longmapsto$ /\textipa{u\tone{55}}/.
\end{center}
The initial letters \textit{y} and \textit{w} do not introduce independent consonantal onsets in these examples. The two syllables are instead factorized as
\begin{center}
    \textit{衣} $\longmapsto$ (\texttt{[EMPTY]}, /\textipa{i}/, /\textipa{\tone{55}}/), \\
    \textit{乌} $\longmapsto$ (\texttt{[EMPTY]}, /\textipa{u}/, /\textipa{\tone{55}}/).
\end{center}

These examples show that Pinyin-to-IPA conversion is \textbf{highly regular but context-conditioned}. Once phrase-level lookup has selected the intended Pinyin reading, deterministic syllable rules can normalize its spelling into an onset, an atomic rime, and a tone. Pinyin therefore serves as the pronunciation-bearing interface that makes IPA-based phonemic factorization applicable to a logographic writing system.

\subsection{Pinyin-to-IPA Correspondences}
\label{sec:zh_pinyin_ipa_mapping}

Unlike Vietnamese Quốc Ngữ, Chinese characters do not provide a systematic graphemic representation of their pronunciation. Phonemic Tokenizer therefore first assigns a Pinyin pronunciation to each character and then converts the selected Pinyin syllable into IPA. Once the Pinyin form has been determined, this conversion is largely systematic, although several letters have context-conditioned realizations.

A Pinyin syllable is analyzed as
\begin{equation}
    q_i \longmapsto (O_i,R_i,T_i),
\end{equation}
where $O_i$, $R_i$, and $T_i$ denote its onset, atomic rime, and lexical tone. The following tables present the normalized mappings used by Phonemic Tokenizer. They represent broad phonological categories rather than narrow phonetic realizations of individual speakers.

\subsubsection{Initial Correspondences}

\begin{table*}[ht]
    \centering
    \resizebox{0.75\linewidth}{!}{
    \begin{tabular}{llll}
    \toprule
    \textbf{Pinyin initial} & \textbf{Normalized IPA} & \textbf{Example} & \textbf{Example IPA} \\
    \midrule
    \textit{b}  & /\textipa{p}/ & \textit{bā} & /\textipa{pa\tone{55}}/ \\
    \textit{p}  & /\textipa{p\super h}/ & \textit{pā} & /\textipa{p\super ha\tone{55}}/ \\
    \textit{m}  & /\textipa{m}/ & \textit{mā} & /\textipa{ma\tone{55}}/ \\
    \textit{f}  & /\textipa{f}/ & \textit{fā} & /\textipa{fa\tone{55}}/ \\
    \midrule
    \textit{d}  & /\textipa{t}/ & \textit{dā} & /\textipa{ta\tone{55}}/ \\
    \textit{t}  & /\textipa{t\super h}/ & \textit{tā} & /\textipa{t\super ha\tone{55}}/ \\
    \textit{n}  & /\textipa{n}/ & \textit{nā} & /\textipa{na\tone{55}}/ \\
    \textit{l}  & /\textipa{l}/ & \textit{lā} & /\textipa{la\tone{55}}/ \\
    \midrule
    \textit{g}  & /\textipa{k}/ & \textit{gā} & /\textipa{ka\tone{55}}/ \\
    \textit{k}  & /\textipa{k\super h}/ & \textit{kā} & /\textipa{k\super ha\tone{55}}/ \\
    \textit{h}  & /\textipa{x}/ & \textit{hā} & /\textipa{xa\tone{55}}/ \\
    \midrule
    \textit{j}  & /\textipa{\t{t\textctc}}/ & \textit{jī} & /\textipa{\t{t\textctc}i\tone{55}}/ \\
    \textit{q}  & /\textipa{\t{t\textctc}\super h}/ & \textit{qī} & /\textipa{\t{t\textctc}\super hi\tone{55}}/ \\
    \textit{x}  & /\textipa{\textctc}/ & \textit{xī} & /\textipa{\textctc i\tone{55}}/ \\
    \midrule
    \textit{zh} & /\textipa{\t{t\:s}}/ & \textit{zhā} & /\textipa{\t{t\:s}a\tone{55}}/ \\
    \textit{ch} & /\textipa{\t{t\:s}\super h}/ & \textit{chā} & /\textipa{\t{t\:s}\super ha\tone{55}}/ \\
    \textit{sh} & /\textipa{\:s}/ & \textit{shā} & /\textipa{\:sa\tone{55}}/ \\
    \textit{r}  & /\textipa{\:z}/ & \textit{ráo} & /\textipa{\:zau\tone{35}}/ \\
    \midrule
    \textit{z}  & /\textipa{\t{ts}}/ & \textit{zā} & /\textipa{\t{ts}a\tone{55}}/ \\
    \textit{c}  & /\textipa{\t{ts}\super h}/ & \textit{cā} & /\textipa{\t{ts}\super ha\tone{55}}/ \\
    \textit{s}  & /\textipa{s}/ & \textit{sā} & /\textipa{sa\tone{55}}/ \\
    \bottomrule
    \end{tabular}}
    \caption{Mapping from Pinyin initials to the normalized IPA onsets used by Phonemic Tokenizer.}
    \label{tab:zh_initial_mapping}
\end{table*}

Pinyin distinguishes 21 consonantal initials. Table \ref{tab:zh_initial_mapping} maps them to their normalized IPA representations. Pinyin letters such as \textit{b}, \textit{d}, and \textit{g} represent voiceless unaspirated stops rather than the voiced stops suggested by their conventional Latin values. Their counterparts \textit{p}, \textit{t}, and \textit{k} are distinguished by aspiration.

A syllable without a consonantal initial is assigned \texttt{[EMPTY]} in its onset slot. The Pinyin letters \textit{y} and \textit{w} at syllable-initial position generally indicate the organization of an onsetless final rather than independent consonantal onsets. For example,
\begin{align}
    \textit{衣}~(\textit{yī}) &\longmapsto \left(\texttt{[EMPTY]}, /\textipa{i}/, /\textipa{\tone{55}}/ \right), \\
    \textit{乌}~(\textit{wū}) &\longmapsto \left(\texttt{[EMPTY]}, /\textipa{u}/, /\textipa{\tone{55}}/ \right).
\end{align}

\subsubsection{Basic Final Correspondences}

\begin{table}[ht]
    \centering
    \small
    \begin{tabular}{lll}
    \toprule
    \textbf{Pinyin final} & \textbf{Normalized IPA} & \textbf{Example} \\
    \midrule
    \textit{a}   & /\textipa{a}/       & \textit{mā} \\
    \textit{o}   & /\textipa{wo}/      & \textit{bō} \\
    \textit{e}   & /\textipa{7}/       & \textit{gē} \\
    \textit{ê}   & /\textipa{E}/       & \textit{ê} \\
    \textit{ai}  & /\textipa{ai}/      & \textit{lái} \\
    \textit{ei}  & /\textipa{ei}/      & \textit{lèi} \\
    \textit{ao}  & /\textipa{au}/      & \textit{hǎo} \\
    \textit{ou}  & /\textipa{ou}/      & \textit{dōu} \\
    \textit{an}  & /\textipa{an}/      & \textit{ān} \\
    \textit{en}  & /\textipa{9n}/      & \textit{ēn} \\
    \textit{ang} & /\textipa{a\ng}/    & \textit{máng} \\
    \textit{eng} & /\textipa{9\ng}/    & \textit{dēng} \\
    \textit{ong} & /\textipa{U\ng}/    & \textit{dōng} \\
    \textit{er}  & /\textipa{a\:r}/    & \textit{ér} \\
    \bottomrule
    \end{tabular}
    \caption{Mapping from basic Pinyin finals to normalized atomic rimes.}
    \label{tab:zh_basic_final_mapping}
\end{table}

Table~\ref{tab:zh_basic_final_mapping} lists finals without a leading high-vowel medial. Each complete final is mapped to one atomic rime, even when it contains a nucleus and a nasal or rhotic coda. The final written as \textit{o} is realized with a labial transition in syllables such as \textit{bō}, \textit{pō}, \textit{mō}, and \textit{fó}. The precise phonetic realization varies, but it is normalized to a single rime category in the tokenizer.

\subsubsection{Finals with \textit{i}, \textit{u}, and \textit{\"u}}

\begin{table*}[ht]
    \centering
    \resizebox{0.65\linewidth}{!}{
    \begin{tabular}{lll}
    \toprule
    \textbf{Pinyin final} & \textbf{Normalized IPA} & \textbf{Example} \\
    \midrule
    \textit{i}    & /\textipa{i}/        & \textit{mǐ} \\
    \textit{ia}   & /\textipa{ia}/       & \textit{jiā} \\
    \textit{ie}   & /\textipa{iE}/       & \textit{xiě} \\
    \textit{iao}  & /\textipa{iau}/      & \textit{xiǎo} \\
    \textit{iou}  & /\textipa{iou}/      & \textit{liú} \\
    \textit{ian}  & /\textipa{iEn}/      & \textit{tiān} \\
    \textit{in}   & /\textipa{in}/       & \textit{xīn} \\
    \textit{iang} & /\textipa{ia\ng}/    & \textit{liàng} \\
    \textit{ing}  & /\textipa{i\ng}/     & \textit{míng} \\
    \textit{iong} & /\textipa{iU\ng}/    & \textit{xiōng} \\
    \bottomrule
    \end{tabular}}
    \caption{Mapping of Pinyin finals containing the \textit{i} medial. The underlying final \textit{iou} is written as \textit{iu} after an initial.}
    \label{tab:zh_i_final_mapping}
\end{table*}

\begin{table}[ht]
    \centering
    \resizebox{0.65\linewidth}{!}{
    \begin{tabular}{llll}
    \toprule
    \textbf{Group} & \textbf{Pinyin final} & \textbf{Normalized IPA} & \textbf{Example} \\
    \midrule
    \multirow{9}{*}{\textit{u}} & \textit{u}    & /\textipa{u}/        & \textit{lù} \\
    & \textit{ua}   & /\textipa{ua}/       & \textit{huā} \\
    & \textit{uo}   & /\textipa{uo}/       & \textit{guó} \\
    & \textit{uai}  & /\textipa{uai}/      & \textit{kuài} \\
    & \textit{uei}  & /\textipa{uei}/      & \textit{duì} \\
    & \textit{uan}  & /\textipa{uan}/      & \textit{guān} \\
    & \textit{uen}  & /\textipa{u9n}/      & \textit{lùn} \\
    & \textit{uang} & /\textipa{ua\ng}/    & \textit{guāng} \\
    & \textit{ueng} & /\textipa{u9\ng}/    & \textit{wēng} \\
    \midrule
    \multirow{4}{*}{\textit{\"u}}
    & \textit{\"u}    & /\textipa{y}/      & \textit{lǜ} \\
    & \textit{\"ue}   & /\textipa{yE}/     & \textit{lüè} \\
    & \textit{\"uan}  & /\textipa{yEn}/    & \textit{juǎn} \\
    & \textit{\"un}   & /\textipa{yn}/     & \textit{jūn} \\
    \bottomrule
    \end{tabular}}
    \caption{Mapping of Pinyin finals containing the \textit{u} and \textit{\"u} medials. The table shows underlying final forms before Pinyin spelling contractions are applied.}
    \label{tab:zh_u_final_mapping}
\end{table}

Tables~\ref{tab:zh_i_final_mapping} and \ref{tab:zh_u_final_mapping} present finals containing the three Mandarin medials conventionally represented by \textit{i}, \textit{u}, and \textit{\"u}. The IPA representations include the medial as part of the atomic rime.

\subsubsection{Context-Conditioned Pinyin Spellings}

Several Pinyin finals have different written forms depending on whether a consonantal initial is present. Phonemic Tokenizer expands these surface forms before converting them into IPA, as summarized in Table~\ref{tab:zh_pinyin_normalization}.

\begin{table*}[ht]
    \centering
    \resizebox{\linewidth}{!}{
    \begin{tabular}{llll}
    \toprule
    \textbf{Surface form} & \textbf{Normalized final} & \textbf{Example} & \textbf{Normalized IPA} \\
    \midrule
    \textit{iu} after an initial & \textit{iou} & \textit{liú} & /\textipa{liou\tone{35}}/ \\
    \textit{ui} after an initial & \textit{uei} & \textit{duì} & /\textipa{tuei\tone{51}}/ \\
    \textit{un} after an initial & \textit{uen} & \textit{lùn} & /\textipa{lu9n\tone{51}}/ \\
    \midrule
    \textit{yi}, \textit{ya}, \textit{ye} & \textit{i}, \textit{ia}, \textit{ie} & \textit{yī}, \textit{yā}, \textit{yè} & /\textipa{i\tone{55}}/, /\textipa{ia\tone{55}}/, /\textipa{iE\tone{51}}/ \\
    \textit{yao}, \textit{you} & \textit{iao}, \textit{iou} & \textit{yào}, \textit{yǒu} & /\textipa{iau\tone{51}}/, /\textipa{iou\tone{214}}/ \\
    \textit{yan}, \textit{yin}, \textit{yang}, \textit{ying} & \textit{ian}, \textit{in}, \textit{iang}, \textit{ing} & \textit{yán}, \textit{yīn}, \textit{yáng}, \textit{yīng} & Corresponding \textit{i}-group rimes \\
    \midrule
    \textit{wu}, \textit{wa}, \textit{wo} & \textit{u}, \textit{ua}, \textit{uo} & \textit{wū}, \textit{wā}, \textit{wǒ} & Corresponding \textit{u}-group rimes \\
    \textit{wai}, \textit{wei}, \textit{wan}, \textit{wen} & \textit{uai}, \textit{uei}, \textit{uan}, \textit{uen} & \textit{wài}, \textit{wèi}, \textit{wǎn}, \textit{wèn} & Corresponding \textit{u}-group rimes \\
    \textit{wang}, \textit{weng} & \textit{uang}, \textit{ueng} & \textit{wáng}, \textit{wēng} & Corresponding \textit{u}-group rimes \\
    \midrule
    \textit{yu}, \textit{yue}, \textit{yuan}, \textit{yun} & \textit{\"u}, \textit{\"ue}, \textit{\"uan}, \textit{\"un} & \textit{yú}, \textit{yuè}, \textit{yuán}, \textit{yún} & Corresponding \textit{\"u}-group rimes \\
    \textit{u} after \textit{j}, \textit{q}, or \textit{x} & \textit{\"u} & \textit{jū}, \textit{qù}, \textit{xū} & /\textipa{y}/ within the corresponding rime \\
    \bottomrule
    \end{tabular}}
    \caption{Context-conditioned Pinyin spellings normalized before IPA conversion. Initial \textit{y} and \textit{w} in these forms are treated as orthographic markers rather than consonantal onset tokens.}
    \label{tab:zh_pinyin_normalization}
\end{table*}

A further context-dependent case concerns the Pinyin final \textit{i}. Following ordinary initials, it represents the high front vowel $/\textipa{i}/$. Following the dental sibilants \textit{z}, \textit{c}, and \textit{s}, or the retroflex initials \textit{zh}, \textit{ch}, \textit{sh}, and \textit{r}, it represents a syllabic apical segment rather than $/\textipa{i}/$. These contexts are assigned separate normalized rime entries:

\begin{equation}
    \begin{aligned}
        \textit{mi} &\longmapsto /\textipa{mi}/,\\
        \{\textit{zi},\textit{ci},\textit{si}\} &\longmapsto \text{dental apical rime},\\ \{\textit{zhi},\textit{chi},\textit{shi},\textit{ri}\} & \longmapsto \text{retroflex apical rime}.
    \end{aligned}
    \label{eq:zh_apical_i}
\end{equation}

This distinction prevents the surface letter \textit{i} from incorrectly collapsing phonologically different Mandarin rimes.

\subsubsection{Tone Correspondences}

Pinyin represents four principal lexical tones using diacritics or the numerical labels 1--4. An unstressed neutral tone is left unmarked or marked with the numeral 5. Table~\ref{tab:zh_tone_mapping} gives the normalized tone mappings.

\begin{table}[ht]
    \centering
    \small
    \begin{tabular}{llll}
    \toprule
    \textbf{Tone} & \textbf{Pinyin mark} & \textbf{Example} & \textbf{Tokenizer value} \\
    \midrule
    First  & Macron       & \textit{mā} & /\textipa{\tone{55}}/ \\
    Second & Acute        & \textit{má} & /\textipa{\tone{35}}/ \\
    Third  & Caron        & \textit{mǎ} & /\textipa{\tone{214}}/ \\
    Fourth & Grave        & \textit{mà} & /\textipa{\tone{51}}/ \\
    Neutral & Unmarked    & \textit{ma} & \texttt{[EMPTY]} \\
    \bottomrule
    \end{tabular}
    \caption{Mapping from Pinyin tone marks to the normalized tone representations used by Phonemic Tokenizer. The neutral tone is assigned \texttt{[EMPTY]} because it has no overt tone mark.}
    \label{tab:zh_tone_mapping}
\end{table}

Taken together, these mappings define the deterministic second stage of the
Chinese tokenization pipeline:
\begin{equation}
    c_i
    \xrightarrow{\psi_{\mathrm{Zh}}}
    q_i
    \xrightarrow{\rho_{\mathrm{Zh}}}
    s_i
    \longmapsto
    (O_i,R_i,T_i),
\end{equation}
where $\psi_{\mathrm{Zh}}$ assigns a Pinyin pronunciation and $\rho_{\mathrm{Zh}}$ converts that pronunciation into normalized IPA. Detailed Pinyin-to-IPA mappings are provided in Tables~\ref{tab:zh_initial_mapping}--\ref{tab:zh_tone_mapping}.

\end{document}